\newif\ifreview
\reviewfalse %

\newif\ifmulticol
\multicoltrue

\ifmulticol
\documentclass[9pt,twocolumn]{extarticle}
\else
\documentclass[9pt]{extarticle}
\usepackage[left=2.5cm, right=2.5cm]{geometry}
\fi

 \fontfamily{ptm}\selectfont

\usepackage[table]{xcolor} %
\usepackage{tikz}
\usetikzlibrary{positioning}
\usepackage[inline]{enumitem}

\DeclareUnicodeCharacter{0301}{\'{e}}

\usepackage{amssymb}
\usepackage{algorithm}
\usepackage{svg}
\usepackage{amsmath}
\usepackage{enumitem}
\usepackage{booktabs}
\usepackage[font=small,labelfont=bf]{caption}
\usepackage[skip=0.5ex]{subcaption}
\usepackage[hidelinks]{hyperref}
\usepackage[capitalize]{cleveref}
\usepackage{authblk}
\usepackage{mathtools}

\usepackage{siunitx}

\usepackage{setspace}
\newenvironment{tight_enumerate}{
\begin{enumerate}
  \setlength{\itemsep}{0pt}
  \setlength{\parskip}{0pt}
}{\end{enumerate}}

\ifreview
\usepackage{bibentry} %
\usepackage{soul} %

\usepackage[switch, pagewise]{lineno} %
\fi

\usepackage{csquotes} 
\usepackage[style=apa,sortcites=true,sorting=nyt,backend=biber, maxbibnames=20]{biblatex}
\DeclareLanguageMapping{american}{american-apa}
\ifreview
\linenumbers %

\newcommand{\R}[1]{\label{#1_}\linelabel{#1}} %

\newcommand\declquotedtext[2]{\expandafter\def\csname quotedtext@#1 \endcsname{#2}} %
\newcommand\defquotedtext[2]{\declquotedtext{#1}{#2}\textcolor{blue}{#2}} %
\newcommand\usequotedtext[1]{``\csname quotedtext@#1 \endcsname''} %
\newcommand\usequotedtextnomark[1]{\csname quotedtext@#1 \endcsname} %

\else
\newcommand\defquotedtext[2]{#2}

\newcommand\R[1]{ }
\fi

\definecolor{lightgray}{gray}{0.96}
\definecolor{winner}{RGB}{220,255,220} %

\begin{document}

\renewcommand\Affilfont{\fontsize{9}{10.8}\itshape}

\title{FastSurfer-CC: A robust, accurate, and comprehensive framework for corpus callosum morphometry}

\author[1]{Clemens Pollak}
\author[1]{Kersten Diers}
\author[1]{Santiago Estrada}
\author[1]{David Kügler}
\author[1,2,3,*]{Martin Reuter}

\affil[1]{AI in Medical Imaging, German Center for Neurodegenerative Diseases (DZNE), Bonn, Germany}
\affil[2]{A.A. Martinos Center for Biomedical Imaging, Massachusetts General Hospital, Boston, MA, USA}
\affil[3]{Department of Radiology, Harvard Medical School, Boston, MA, USA}
\affil[*]{Corresponding author: Martin.Reuter@dzne.de}

\date{}

\maketitle

\begin{abstract}
\noindent The corpus callosum, the largest commissural structure in the human brain, is a central focus in research on aging and neurological diseases. It is also a critical target for interventions such as deep brain stimulation and serves as an important biomarker in clinical trials, including those investigating remyelination therapies. Despite extensive research on corpus callosum segmentation, few publicly available tools provide a comprehensive and automated analysis pipeline. To address this gap, we present FastSurfer-CC, an efficient and fully automated framework for corpus callosum morphometry. FastSurfer-CC automatically identifies mid-sagittal slices, segments the corpus callosum and fornix, localizes the anterior and posterior commissures to standardize head positioning, generates thickness profiles and subdivisions, and extracts eight shape metrics for statistical analysis. We demonstrate that FastSurfer-CC outperforms existing specialized tools across the individual tasks. Moreover, our method reveals statistically significant differences between Huntington's disease patients and healthy controls that are not detected by the current state-of-the-art.
\end{abstract}


\begin{figure}[tb]
    \ifmulticol
    \includegraphics[width=1.05\linewidth]{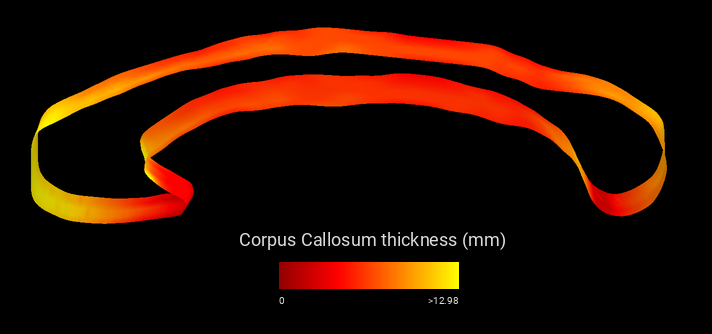}
    \else
    \includegraphics[width=0.6\linewidth]{figures/cc_mesh_snap.png}
    \centering
    \fi
    \caption{Thickness of an example corpus callosum, shown on a 3D surface model using WhipperSnapPy~\autocite{whippersnappy} for visualization.}
    \label{fig:3d_model}
\end{figure}

\section{Introduction}

The corpus callosum and fornix, as well as anterior and posterior commissures, are white matter bundles central to the communication between hemispheres, memory recall tasks, and olfaction.
In particular, the corpus callosum is the largest brain commissure and is associated with many diseases, for example, epilepsy~\autocite{unterberger2016corpus}, autism~\autocite{piven1997mri}, schizophrenia~\autocite{woodruff1995meta}, multiple sclerosis~\autocite{ozturk2010mri}, cerebral palsy~\autocite{jaatela2023altered}, Parkinson's disease~\autocite{yang2023white}, bipolar disorder~\autocite{grande2016bipolar}, and Alzheimer's disease~\autocite{di2010vivo}.
The corpus callosum (CC) is an extensively studied structure, which is presented with high contrast in structural MR imaging and shows complex anatomical changes in healthy aging and disease~\autocite{luders2010development}. 
Adjacent to the CC lies the body of the fornix (FN), which is part of the limbic system and does not connect the hemispheres directly but forms two C-shaped arches that connect to the hippocampus on the left and right hemispheres and merge below the corpus callosum. %
The fornix is most prominently involved in Alzheimer's disease~\autocite{lacalle2023fornix, nowrangi2015fornix} and mild cognitive impairment~\autocite{nowrangi2015fornix, zhuang2012abnormalities}. %
\begin{figure}[t]
    \centering
    \begin{subfigure}[a]{1\linewidth}
    \ifmulticol
    \includegraphics[width=1\linewidth]{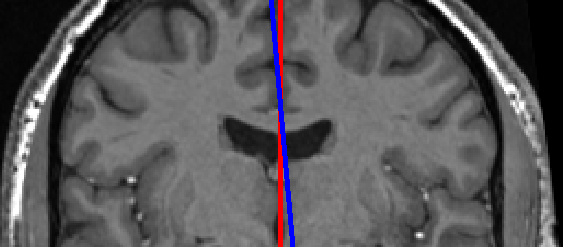}
    \else
    \centering
    \includegraphics[width=0.45\linewidth]{figures/planes2.png}
    \fi
    \caption{Two mid-sagittal plane estimates (red \& blue) differ only by a rotation of \SI{5}{\degree} in the coronal plane.}
    \label{fig:midplane_teaser_a}
    \end{subfigure}
    \begin{subfigure}[b]{1\linewidth}
    \ifmulticol
    \includegraphics[width=1\linewidth]{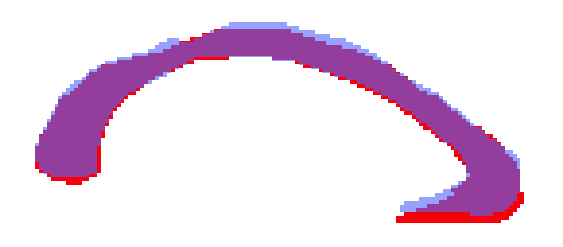}
    \else
    \centering
    \includegraphics[width=0.5\linewidth]{figures/overlaid_cc.png}
    \fi
    \caption{The red mask results from accurate segmentation using the red plane as mid-sagittal plane, while the blue mask results from accurate segmentation using the blue plane.} %
    \label{fig:midplane_teaser_b}
    \end{subfigure}
    \caption{Choice of mid-sagittal plane affects the shape and thickness of corpus callosum segmentation.}
    \label{fig:midplane_teaser}
\end{figure}

The anterior commissure (AC) connects the orbitofrontal, temporal, parietal, and occipital lobes, as well as the insular and entorhinal cortices, with the olfactory bulbs, the septal area, and the amygdalae~\autocite{pinto2017split, raybaud2010corpus, ccavdar2021complex}.
The posterior commissure (PC) has been identified as the connection between pre- and postcentral gyri, the superior parietal region in the left hemisphere to the temporal region, and the lateral occipital and superior parietal regions of the contra-lateral hemisphere~\autocite{pinto2017split}, and is also connected to thalamic nuclei, superior colliculus, and the habenular nuclei at its origins~\autocite{pinto2017split}.
Despite their wide reach, the AC \& PC present themselves as distinct white matter tracts almost completely surrounded by gray matter in the mid-sagittal view.
Although the function of AC \& PC is not well understood, their contrast to the surrounding gray matter has made them essential neuroanatomical landmarks used for localization of neuroanatomical structures during deep brain stimulation (DBS), stereotactic and functional surgery~\autocite{baudo2022three}, the definition of corpus callosum sub-segments~\autocite{jancke1997relationship}, as well as the standardization of head pose for automated morphometric analysis~\autocite{adamson2014software}. Their relative locations -- with respect to other structures and each other -- have also been used as a biomarker~\autocite{vermeulen2023morphological}.

\begin{figure*}[t]
    \includegraphics[width=1\linewidth, trim={0cm 0cm 4.3cm 0cm}, clip]{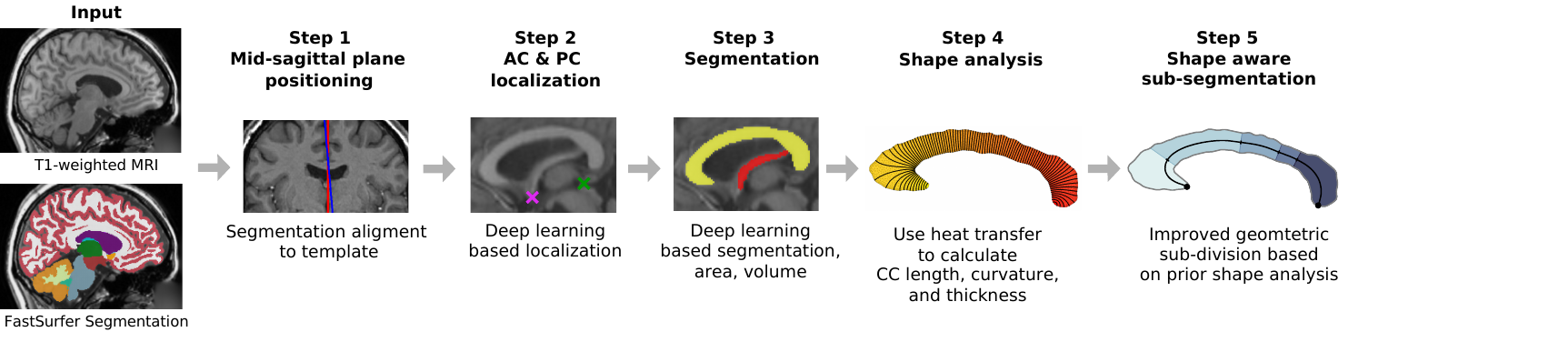}
    \caption{Overview of the proposed pipeline for corpus callosum morphometry.}
    \label{fig:pipeline_overview}
\end{figure*}

\subsection{Comprehensive integrated approach}

Researchers who investigate corpus callosum changes in aging, disease, and intervention studies require an accurate tool for deriving markers of corpus callosum thickness, shape, area, and volume. Instead of focusing on a single task, our framework \mbox{FastSurfer-CC} integrates the following five sub-tasks into a comprehensive pipeline (see Figure \ref{fig:pipeline_overview}):
\begin{tight_enumerate}
    \item Find the mid-sagittal plane,
    \item localize the anterior and posterior commissures,
    \item segment CC and fornix,
    \item derive morphometric estimates for CC area, length, curvature, and thickness, and
    \item sub-divide the CC with a meaningful geometric sub-segmentation method.
\end{tight_enumerate}
Finding an accurate mid-sagittal plane is motivated by the lack of clear CC boundaries in the lateral direction, often addressed by segmenting the CC in mid-sagittal slices of the brain within a pre-defined width~\autocite{adamson2014software, ardekani2013automated, fischl2012freesurfer}. Determining the mid-sagittal slices accurately and independently of the image slicing direction is essential, as the shape, thickness, and area of the corpus callosum can differ considerably depending on the position and orientation of the selected mid-sagittal plane (see Figure~\ref{fig:midplane_teaser}). %
The second step, anterior- and posterior commissure localization, is required for multiple tasks: standardizing the head position in the remaining direction (nodding), finding the corpus callosum end-points, and estimating the CC location in the mid-sagittal slice as initialization for the segmentation task. Following the segmentation, the analysis can be run independently of the images' voxel grid by converting the segmentation mask into a triangle mesh with sub-voxel resolution along the boundary. In the fourth step, building on the geometric representation of the CC, shape summary metrics like area, principal length, curvature, and local thickness can be calculated reliably. Finally, a novel and improved geometric sub-segmentation is enabled that adapts to the curved shape of the CC naturally.

\subsection{Related work}

Existing literature frequently reduces the challenge of CC morphometry to one of the aforementioned tasks, without integrating or evaluating the effects of other pipeline steps~\autocite{jlassi20243dcc, chandra2023pccs, gajawelli2024surface}. We also target superior accuracy for each individual component and therefore benchmark components in isolation against specialized tools. However, jointly implementing and validating our novel and comprehensive framework (shown in Figure~\ref{fig:pipeline_overview}) avoids incompatibilities, e.g., a segmentation tool only working on incorrectly or poorly chosen mid-sagittal planes; accurate thickness estimation, but on low-quality segmentation maps; or chaining tools with large computational overhead due to inconsistent pre- and postprocessing steps in the pipeline.

\subsubsection{Mid-sagittal plane positioning}

Finding a mid-sagittal plane for corpus callosum segmentation can be based on a template, global or local symmetry, or anatomical landmarks.
Sometimes the selection of mid-sagittal slices from the image (without re-slicing) has also been formulated as a classification task. This has been achieved manually~\autocite{platten2020mri} and with a convolutional neural network (CNN) classifier~\autocite{brusini2022automatic}. Selecting existing slices from the image, however, oversimplifies the problem, as it disregards the left-right head rotation and tilting and might cause biases in downstream analysis. For example, tighter padding as applied in tremor cases, e.g., from MS, causes the head to be more upright compared to controls, where the head positioning is less restrained, eventually leading to bias in group comparisons as outlined in Figure~\ref{fig:midplane_teaser}. To correct for these effects, rotating and re-slicing of the image is necessary. Registration-based approaches, e.g., to MNI space~\autocite{huang2021deformation, adamson2014software, gajawelli2024surface} can standardize translation and rotation, thereby finding a mid-sagittal plane from the atlas space. %
FreeSurfer's~\autocite{fischl2012freesurfer} \textit{mri\_cc} tool uses the symmetry of the whole brain segmentation as an initial estimate. It employs multiple heuristics to find the mid-sagittal plane, among them an optimization step rotating the plane and maximizing an alignment score based on the left and right cerebral white matter. The \textit{Yuki} software package~\autocite{ardekani2022new, ardekani2013automated} also relies on registration but refines the mid-sagittal plane based on 8 landmarks, while \textit{CCSegThickness} (short CCSeg, \cite{adamson2014software}) incorporates a registration with FLIRT~\autocite{jenkinson2012fsl} to an MNI template. %

Thus far, a joint evaluation and comparison of these methods is missing, %
and it is currently unclear which general approach is superior for corpus callosum segmentation. We aim to close this gap in our study with a rater comparison.

\subsubsection{Corpus callosum and fornix segmentation}

The two predominant approaches for medical image segmentation are deep learning and registration methods.
For segmentation of the CC, atlas-based approaches %
are employed by Yuki~\autocite{ardekani2022new, ardekani2013automated} and CCSeg~\autocite{adamson2014software}. These methods often require manual cleanup \autocite{adamson2014software, raaf2023hand} due to ``misclassifications of pericallosal veins or the fornix''\autocite{raaf2023hand}. FreeSurfer's \textit{mri\_cc} determines an initial estimate based on the white matter of a pre-existing whole brain segmentation and then uses thresholding and other post-processing steps to segment the corpus callosum. Van Schependom and colleagues~\autocite{van2016reliability} proposed an active shape model combined with an atlas registration for an initial estimate. Unfortunately, the method is not easily reproducible with no released implementation or technical documentation.\R{R:method_selection} \defquotedtext{Q:method_selection}{In this work we will compare to existing methods that are accessible, interoperable, and reusable~\autocite{chue_hong_2022_fair4rs}, enabling us to isolate method components for fair comparison.}
\R{R:mricloud}\defquotedtext{Q:mricloud}{MRICloud~\autocite{mori2016mricloud} is an exclusively cloud based processing tool that includes atlas based corpus callosum segmentation. Several dataset protection regulations and data usage agreements, however, do not permit uploading medical data to a cloud service.}
Multiple deep learning-based segmentation methods have been proposed. Among them are a classical U‐Net~\autocite{platten2020mri}, a modified residual attention U-Net~\autocite{chandra2023pccs}, a combination of Bi-Directional Convolutional LSTMs with a U-Net~\autocite{wong2023brain} for multi-modal segmentation, and a probabilistic neural network~\autocite{jlassi20243dcc}. Generally, machine learning tools have been found to be more accurate in corpus callosum segmentation~\autocite{cover2018computational}, however, the present methods were not compared with established tools, and trained models are not available, which greatly reduces usability. To close this gap, we contribute a novel, robust, well-validated, open-source CC segmentation method.

For the segmentation of the fornix, similar registration-based~\autocite{chang2022open, fischl2012freesurfer} and deep learning tools~\autocite{greve2021deep} exist. FreeSurfer~\autocite{fischl2012freesurfer} contains two tools for fornix segmentation. \textit{mri\_cc} performs joint, corpus callosum and fornix segmentation in the mid-sagittal plane, while \textit{ScLimbic}~\autocite{greve2021deep} segments 5 structures of the limbic system (hypothalamus, nucleus accumbens, fornix, basal forebrain, and septal nuclei) using a standard U-Net. ScLimbic segments the whole fornix and is not limited to the mid-sagittal slices -- allowing for segmentation of the anterior pillars and the fimbria. The same is true for the registration-based Open-Source Hypothalamic-ForniX (OSHy-X) Atlases and Segmentation Tool~\autocite{chang2022open}.

\begin{figure*}[!ht]
    \centering
    \ifmulticol
    \includegraphics[width=17cm, trim={0.5cm 1cm 1cm 0.5cm},clip]{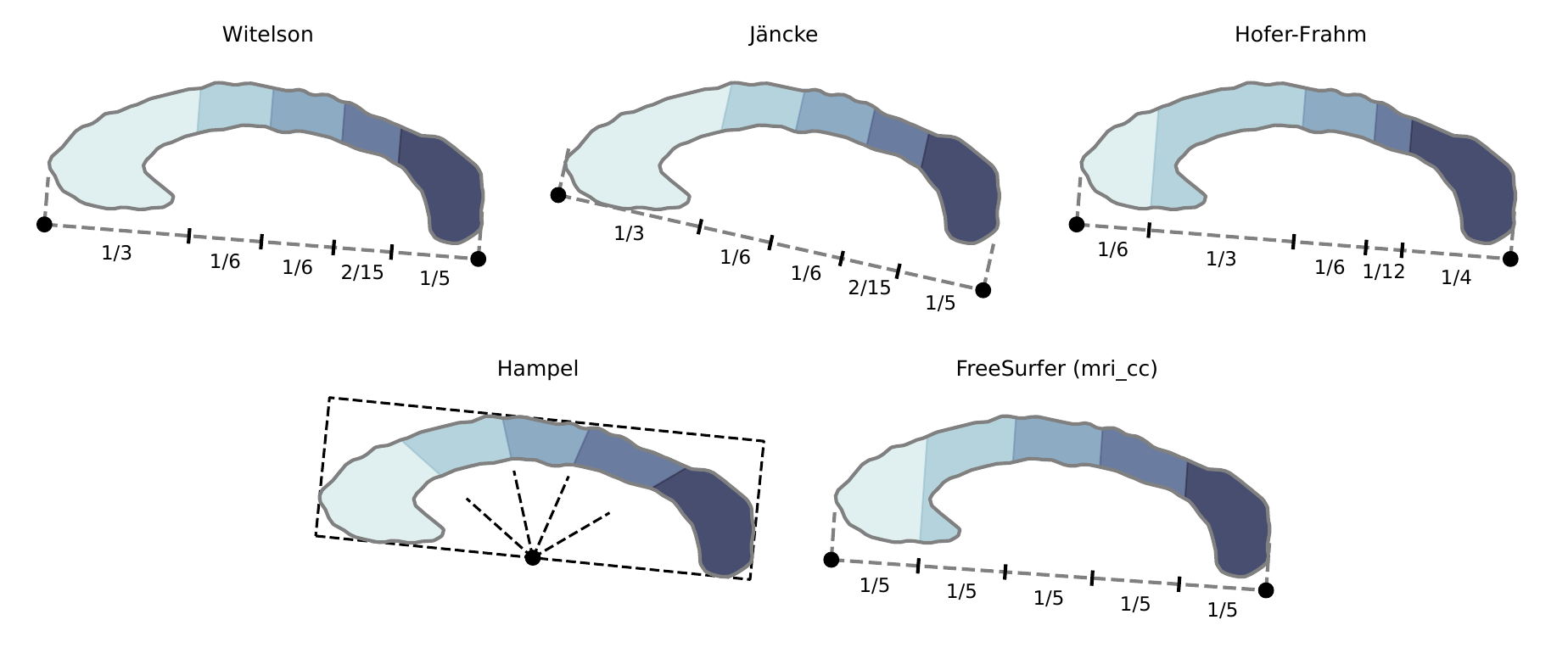}
    \else
    \includegraphics[width=1\linewidth, trim={0.5cm 1cm 1cm 0.5cm},clip]{figures/cc_subdivisions_figure_fs_aligned.pdf}
    \fi
    \caption{Overview of previously proposed sub-division schemes. To the left is the anatomical anterior.}
    \label{fig:subsegmentation_comparison}
\end{figure*}

\subsubsection{Corpus callosum morphometry}

While the area of the corpus callosum cross-section is already a valuable biomarker, atrophy in specific regions is associated with different diseases and of broad research interest. Therefore, areas of local sub-segments and thickness across the corpus callosum are highly relevant morphometrics.
A plethora of sub-division schemes have been proposed, some of them based on extensive in-vivo and ex-vivo analysis. Friedrich and colleagues~\autocite{friedrich2020mapping} give an overview of the different approaches for sub-division. The most practical methods for structural MRI are geometric approaches, which allow for a robust and generalizing sub-division. Even when considering only geometric sub-divisions applicable to structural MRI, a number of schemes have been proposed (illustrated in Figure~\ref{fig:subsegmentation_comparison}): 
\begin{enumerate}
    \item The Wittelson scheme~\autocite{witelson1985brain} defines geometric sub-divisions based on lines orthogonal to an \textit{anchor line} through the most anterior and posterior points of the CC.
    \item \cite{jancke1997relationship} improve Wittelsons sub-division, by drawing the anchor line through the anterior and posterior commissures, which is more robust to segmentation errors.
    \item The Hofer-Frahm scheme~\autocite{hofer2006topography} also expands on the Witelson scheme and re-defines the widths of each segment according to findings from diffusion MRI (dMRI) studies (while keeping the same anchor line).
    \item \cite{hampel1998corpus} define an independent sub-division scheme based on equally spaced rays originating from a midpoint on the inferior border of a rectangle fitted around the CC, which yields sub-divisions roughly orthogonal to the CC direction. 
    \item FreeSurfer ({\it mri\_cc})~\autocite{fischl2012freesurfer} uses equally spaced divisions along the primary eigendirection.
\end{enumerate}
Most of the proposed schemes sub-divide the CC along the inferior-superior direction. If the anterior segments are small, this can lead to unintended merging of the rostrum and body of the CC, as seen for \textit{mri\_cc} and Hofer-Frahm in Figure~\ref{fig:subsegmentation_comparison}. To address this shortcoming, while preserving well-motivated sub-division fractions, we propose a shape-aware sub-division method. This method, similar to \cite{hampel1998corpus}, cuts the sub-segments perpendicular to the CC principal direction. %

Beyond area-based analysis of the corpus callosum, prior works have proposed the measurements of overall thickness~\autocite{platten2020mri} and thickness profiles~\autocite{adamson2011thickness, adamson2014software, lee2014application,  fraize2023mapping, van2016reliability}, which can be complemented with shape information~\autocite{ardekani2013automated, van2016reliability, huang2021deformation, joshi2013statistical} and sometimes also surface statistics~\autocite{gajawelli2024surface}. Summary metrics, like mid-sagittal CC area, circularity~\autocite{ardekani2013automated, van2016reliability}, bending angle~\autocite{platten2020mri}, and the corpus callosum index \autocite{figueira2007corpus} have also been shown to be discriminative~\autocite{fujimori2020measurements, platten2020mri, ardekani2013automated, van2016reliability}, reliable and repeatable~\autocite{van2016reliability} and interpretable clinical markers.

\subsection{Contributions}

In this work we contribute to and combine each of the previously discussed sub-tasks of i) head-pose standardization for mid-sagittal plane positioning, ii) CC \& FN segmentation, iii) down-stream morphometry, and iv) sub-division. We provide a comprehensive open-source framework (FastSurfer-CC, see Figure~\ref{fig:pipeline_overview}) as a fast, robust, accurate, and comprehensive tool for advanced corpus callosum morphometry. FastSurfer-CC:
\begin{tight_enumerate}
\item Accurately identifies the optimal mid-sagittal plane tailored for corpus callosum analysis.
\item Outperforms existing specialized tools in localizing the anterior- and posterior commissures, as well as segmenting the corpus callosum and fornix. 
\item Demonstrates superior robustness and accuracy on challenging cases compared to other methods. 
\item Introduces novel, reliable metrics for corpus callosum length and curvature.
\item Provides efficient, precise, and robust thickness estimation along the corpus callosum. 
\item Implements a new shape-aware sub-division approach compatible with previously established schemes.
\end{tight_enumerate}

\section{Methods}

In the following section, we introduce the datasets and present the individual components: mid-sagittal plane positioning, AC \& PC localization, CC \& fornix segmentation, thickness, curvature and length estimation, and CC sub-segmentation (see Figure~\ref{fig:pipeline_overview}). Finally, we describe the evaluation criteria to assess the accuracy and robustness of the aforementioned components.

\subsection{Datasets}

We assemble two types of datasets: 1.\ the training, validation, and test datasets with manual labels, and 2.\ a downstream application dataset without manual labels.

Diverse data for training and testing is critical for method generalization and usability. Here, we start from the FastSurfer training dataset compiled from 12 datasets as described by \cite{henschel2022fastsurfervinn} and add 7T data from the group of MR physics at the German Center for Neurodegenerative Diseases (DZNE) and cases with resection cavities from the Uniklinikum Bonn (UK-Bonn) to further increase heterogeneity. From this large corpus of structural MRI, we select a subset of 280 T1w scans for manual annotation and comparison of mid-sagittal planes. After the labeling process and quality control, we end up with 173 labeled volumes. These are then split into training (93), validation (31), and test-set (30), and an additional test-set consisting of cases that were challenging for raters (19), which includes cases with motion artifacts, low contrast, and brain lesions. In Appendix Figure~\ref{fig:difficult_examples} we show examples of challenging cases, and in Appendix Table~\ref{tab:split} we present a breakdown of the dataset split and provide demographics and scanner data in Appendix Table~\ref{tab:dev_sets_metadata}. 
We evaluate performance on both randomly selected and specifically challenging cases for the following reasons:\begin{tight_enumerate}
    \item Researchers often require methods to capture disease effects. Diseased groups, however, can show strong atrophy or other anomalies that make processing more challenging.
    \item Datasets of specific diseases can under-represent the general population variance because of study exclusion criteria or selection bias.
    \item A thorough limitation analysis helps define the constraints in which software should be used.
\end{tight_enumerate} 
\R{R:PREDICT_HD_description}\defquotedtext{Q:PREDICT_HD_description}{
Additionally, we apply our method to a subset of 1268 scans of the PREDICT-HD dataset (predict Huntington's disease, \cite{paulsen2008detection}) as a downstream evaluation. We show an overview of the dataset demographics and clinical variables in Appendix Table~\ref{tab:appendix_predicthd}. For this dataset we analyze group differences between Huntington's disease patients (N=992) and healthy controls (N=276) using the proposed morphometrics.}
For each participant, \R{R:HD_visit}we select their third visit and the best quality T1w image of that visit. We also test whether the down-stream metrics produced by (only) our method show group differences.
\R{R:ICC_method_dataset}\defquotedtext{Q:ICC_method_dataset}{Finally, we evaluate the reliability of our method by selecting cases from the previous group analysis with at least two T1w scans in the same session (resulting in 389 scans). For these scans morphological changes are expected to be minimal, but they differ, for example, in quality and head position. We can then assess reliability by testing whether our method can produce consistent morphometrics across the two scans.}
The studies were approved by the ethics board of the respective institutions.

\subsection{Mid-sagittal plane positioning}
\label{sec:midplane_finding}

We find the mid-sagittal plane by registering FastSurfer's whole brain segmentation with the segmentation of the \textit{fsaverage} template. More specifically, we generate two point-clouds with the centroid points of all labels present in both segmentation maps. Following this, we rigidly register the point-clouds using singular value decomposition, which provides the homogeneous transformation matrix $T$, that maps the case segmentation space to the fsaverage template.
This template is based on 40 cases from Washington University, collected by Randy Buckner and colleagues~\autocite{Buckner40ADNI60testing, fsaverageWiki}. It is also roughly aligned to the MNI305~\autocite{evans19933d} template. The template's mid-sagittal plane $P_{fsaverage}$ lies exactly on the middle slice in the L/R direction. Therefore, we can determine the mid-sagittal plane of the original volume $P_{target}$ by mapping $P_{fsaverage}$ to the original images space: $P_{target} = T^{-1} \times P_{fsaverage}$.

\subsection{Anterior- and posterior commissure localization}

Since the mid-sagittal plane usually intersects the AC \& PC points, we use the transformation $T$ from the registration step to extract a mid-sagittal slice from the original volume for localization -- effectively constraining the task to a 2D localization. 
To tackle this task, we train a classical DenseNet~\autocite{huang2016densely}, that predicts two floating-point numbers as outputs and mean squared error as loss. These outputs determine the position of AC \& PC in the image. We regularize image dimensions by resampling all images to \SI{1}{\milli\meter} resolution and cropping the input slices to a size of $64\times64$, initially centered around the third ventricle. We augment the network training by shifting the field of view, making random contrast adjustments, adding Gaussian noise, and randomly scaling intensities. We choose the AdamW optimizer with gradient clipping.
We further aim to increase the model's accuracy and robustness using four additional strategies: i) We increase dataset variance and size by introducing a pre-training step using the FastSurfer training set~\autocite{henschel2022fastsurfervinn} with labels generated by the acpc\_detect tool. ii) We mitigate out-of-distribution failures in case of inaccurate mid-sagittal plane positioning by training the model also on slices adjacent to the estimated optimal mid-sagittal slice. iii) We mitigate out-of-distribution errors for inaccurate fields-of-view by using an iterative inference strategy that refines the field-of-view according to the network predictions (and then runs the network again). Finally, iv) a total of 5 slices (the mid-sagittal slice plus two to either side) are used to estimate AC \& PC locations and averaged.

\begin{figure}[t]
    \centering
    \ifmulticol
    \includegraphics[width=1\linewidth, trim={0.3cm 0cm 0cm 0cm},clip]{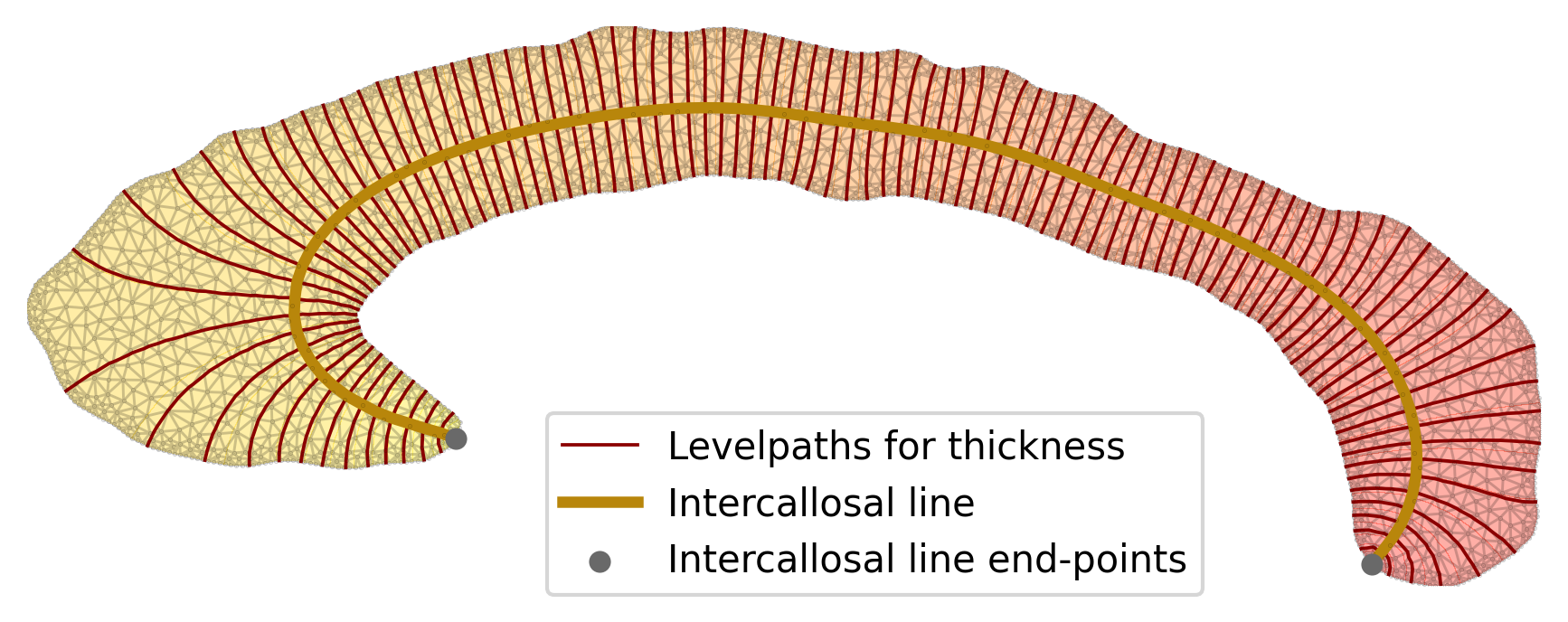}
    \else
    \includegraphics[width=0.5\linewidth, trim={0.3cm 0cm 0cm 0cm},clip]{figures/levelsets2.png}
    \fi
    \caption{Final state of thickness estimation, where the intercallosal line and thickness levelpaths are calculated on the corpus callosum mesh. The solution to the Laplace equation is shown as a gradient from yellow to red.}
    \label{fig:levelsets}
\end{figure}

\subsection{Segmentation}

Similar to AC/PC localization, we segment the CC on the mid-sagittal and adjacent slices. Here, we always consider enough slices to cover at least \SI{2.5}{\milli\meter} to the left and right of the mid-sagittal plane, resulting in a standardized volumetric estimate. %
To aid clear separation of corpus callosum and fornix, we also segment the fornix in these mid-sagittal slices, %
resulting in CC and fornix outputs similar to FreeSurfer's \textit{mri\_cc}. However, while FreeSurfer always segments and computes volumes based on only 5 slices, we increase the number of slices for sub-millimeter voxels. \R{R:cc_volume}\defquotedtext{Q:cc_volume}{We then weigh the volume contribution of the first and last slice appropriately and report a corrected volume estimate that reflects a consistent width of \SI{5}{\milli\meter} independent of the voxel resolution.}
For this segmentation task, we train a variant of the FastSurferVINN~\autocite{henschel2022fastsurfervinn} architecture. This variant only operates on sagittal slices but retains the multi-slice input (i.e., providing neighboring slices as channels) and the VINN layers (Voxel-size Independent Neural Network layers, for interpolation of the latent space). We train the network with Dice and cross-entropy loss. For data augmentation we vary the cropping of the field of view and perform random contrast changes. %
We use the SGD optimizer and cosine annealing with warm restarts as the learning rate scheduler.

\begin{figure}[t]
    \centering
    \ifmulticol
    \includegraphics[width=1\linewidth, trim={1.7cm 0.4cm 1.5cm 0.5cm},clip]{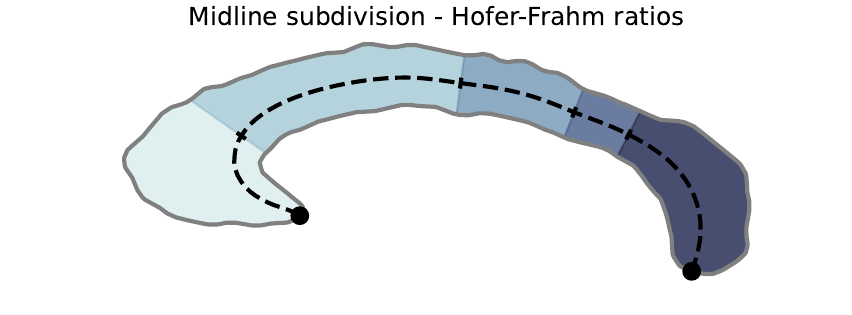}
    \else
    \includegraphics[width=0.5\linewidth, trim={1.7cm 0.4cm 1.5cm 0.5cm},clip]{figures/our_segmentation.pdf}
    \fi
    \caption{Proposed sub-segmentation strategy with division perpendicular to the intercallosal line. Here we use the ratios of the Hofer-Frahm sub-segmentation scheme, but other ratios are also possible. For comparison, the same corpus callosum is used as in Figure~2.}
    \label{fig:our_subsegmentation}
\end{figure}

\subsection{Thickness, curvature and length estimation}

Similar to CCSeg~\autocite{adamson2011thickness}, we aim to provide a method for localized morphometrics to help identify CC regions with significant thickness differences. Therefore, we develop a method to measure CC thickness with arbitrary sampling density (see Figure~\ref{fig:levelsets} for an example with 100 thickness measurements). 
We implement four changes to previous solutions: i) shape analysis with triangle meshes instead of on the voxel grid for more accurate sub-voxel shape representation, ii) CC endpoint localization based on our reliable AC/PC landmarks even for unusual CC shapes, iii) thickness estimates via the solution to the Laplace equation (boundary conditions feature a positive and a negative charge at the opposing inferior and superior boundaries), and iv) efficient computation by estimating the level sets of the rotated Laplace gradients instead of tracing the gradient curves.
Note that the zero level set to the original Laplace solution connects the CC end-points. This \textit{intercallosal} line reliably follows the bend of the CC and can therefore be used to calculate CC curvature, length, and support different geometry-aware sub-division schemes. Level sets to the rotated solution (levelpaths for thickness in Figure~\ref{fig:levelsets}) can be employed to similarly estimate the thickness profiles at arbitrary sampling points along the intercallosal line. We provide a detailed description of our method in Appendix~\ref{appendix:laplace}.

\subsection{Sub-segmentation}

While most previously proposed sub-segmentation schemes are carefully derived based on ex-vivo and in-vivo analysis of corpus callosum morphology and connectivity, their geometric definitions may not always produce sub-segments consistent with the intended parcellation. \R{R:correction1}\defquotedtext{Q:correction1}{In Figure~\ref{fig:subsegmentation_comparison}, for example, we can observe that the Hofer-Frahm sub-segmentation scheme does not separate CC genu and body as originally intended~\autocite{hofer2006topography}, instead merging the rostrum and anterior CC body into one sub-region.} Other schemes show similar shortcomings due to the suboptimal slicing angles with respect to the CC shape. %

To address the limitations of previous CC sub-segmentation schemes, we propose a novel and robust approach for geometric sub-segmentation: we divide the CC perpendicular to the intercallosal line. In Figure~\ref{fig:our_subsegmentation}, we apply the anatomically motivated fractions from the Hofer-Frahm scheme along the principal direction of the CC shape. This approach leads to the intended division between the posterior and superior genu.
We provide a customizable, parametric implementation for all the geometric sub-segmentation schemes shown in this paper, which allows researchers to redefine sub-segmentation -- e.g.\ according to thickness changes derived by our tool.

\subsection{Implementation details}

The FastSurfer-CC framework combines all of the previously discussed method steps and combines them into an efficient pipeline with the following computational improvements for speed and accuracy:
\begin{enumerate}
    \item We interpolate only required slices into the mid-sagittal space for AC \& PC localization and CC segmentation.
    \item We initialize the iterative AC \& PC localization using the third ventricle label.
    \item We use the final AC \& PC locations to crop a patch based on the AC \& PC localization for the segmentation network, then we map the soft-labels (probability maps) and map them back to the original space before thresholding.
    \item We calculate CC circularity and the CC index, which are previously validated CC shape descriptors (see Appendix~\ref{sec:additional_metrics}).
    \item We write outputs asynchronously during processing. %
\end{enumerate}
The pipeline's outputs can then be used for various tasks, including i) volumetric analysis of the corpus callosum (see Figure~\ref{fig:3d_model}),
ii) inclusion of corpus callosum metrics in statistical analysis, and
iii) standardization of head pose and position (AC at the origin, PC placed on the anterior-posterior axis and the CC centered on the left-right axis). All outputs are calculated in less than 10 seconds (Intel Xeon W-2245, 64 GB RAM, Nvidia Titan Xp, solid state drive, $256^3$ volume) or 40 seconds on a MacBook Pro (2.3 GHz Quad-Core Intel Core i7, 16 GB RAM, solid state drive, same volume) via docker containerization -- despite the lack of a dedicated GPU.

\vspace{3cm}

\subsection{Validation metrics}

\subsubsection{Manual comparison of mid-sagittal plane candidates}

With the goal of identifying the best method for mid-sagittal plane positioning, we designed a custom rating tool based on Freeview -- FreeSurfer’s visualization user interface~\autocite{fischl2012freesurfer}. Our tool converts a homogeneous transformation into a plane (see Section~\ref{sec:midplane_finding}) that can be displayed independently of the volume's voxel grid. This enables assessing whether the plane cuts the corpus callosum along its thickest lateral point while preserving local symmetry -- especially the anterior and posterior commissures.

In a direct method comparison, two blinded experts compared forty mid-sagittal planes generated by each method. For each comparison, we selected twenty challenging plane pairs based on the lowest method agreement and twenty pairs at random. Method agreement was estimated by the average distance between planes within the brain volume. We approximated this using a cylinder centered at the RAS coordinate origin and oriented along the two planes' average normal vector. The difference between planes was then defined as the volume enclosed between them within the cylinder.

\R{R:midplane_comp_explanation}\defquotedtext{Q:midplane_comp_explanation}{Raters were asked to choose the higher quality plane according to the aforementioned criteria. When no method was preferred by a rater, no vote was cast. We report the number of votes per method for each rater individually in the Result Section~\ref{sec:midplane_results}. }

\subsubsection{Localization comparison}

We compare estimates of AC \& PC location with the manually marked locations using the Euclidean distance between prediction and reference standard. Since we pay increased attention to method robustness, we show both mean and median distances. Opposed to the mean, the median is not sensitive to outliers, and the difference between the two measurements can therefore indicate whether a method produces large errors in some cases.

\begin{table*}[ht!]
\centering

\setlength{\tabcolsep}{10pt}

\rowcolors{2}{lightgray}{white}

\begin{tabular}{|l||c|c||c|c|}
\hline
& \multicolumn{2}{c||}{\textbf{Rater 1}} & \multicolumn{2}{c|}{\textbf{Rater 2}} \\
\hline
\textbf{SOTA Method} & \textbf{SOTA} & \textbf{FS-CC} & \textbf{SOTA} & \textbf{FS-CC} \\
\hline
\textbf{robust\_register} 
    & 7 & \cellcolor{winner}\textbf{12} 
    & 3 & \cellcolor{winner}\textbf{13} \\
\textbf{mri\_cc} 
    & \cellcolor{winner}\textbf{6} & 4 
    & 10 & \cellcolor{winner}\textbf{12} \\
\textbf{Yuki} 
    & 4 & \cellcolor{winner}\textbf{12} 
    & 10 & \cellcolor{winner}\textbf{22} \\
\textbf{CCSeg} 
    & 7 & \cellcolor{winner}\textbf{15} 
    & 12 & \cellcolor{winner}\textbf{22} \\
\hline
\end{tabular}

\vspace{0.5em}
\caption{Rater comparisons of the proposed method for positioning the mid-sagittal plane (FS-CC) against four state-of-the-art (SOTA) methods. The table shows votes from two independent raters in a direct comparison. The preferred method for each rater is highlighted in green.}
\label{tab:midplane_rating}
\end{table*}

\subsubsection{Segmentation quality}

To evaluate segmentation performance with respect to the manual reference standard, we use the Dice Similarity Coefficient (DSC) and the Hausdorff Distance (HD). The DSC is defined as
\begin{equation}
\mathit{DSC}(X,Y) = \frac{2\ |X \cap Y|}{|X| + |Y|} \hspace{0.3cm} .
\end{equation}
This metric quantifies the overlap between two binary masks $X$ and $Y$, which in our case represent the location and extent of brain structures. A DSC of zero indicates no overlap between the prediction and the reference standard, while a perfect match corresponds to a DSC of one.
The HD is defined as 
\begin{equation}
\mathit{HD}(X,Y) =   \max\left\{\,\sup_{x \in X} d(x,Y),\ \sup_{y \in Y} d(X,y) \,\right\} .
\end{equation}
\R{R:correction2}\defquotedtext{Q:correction2}{It measures the distance between the boundaries of the binary masks $X$ and $Y$.} To provide a more robust evaluation, we use the 95th percentile of the distance rather than the maximum possible value. The HD is reported in millimeters, where 0 mm signifies a perfect match of the structures up to the 95th percentile. 
To determine whether one method significantly outperforms another, we apply the Wilcoxon signed-rank test, as implemented in the SciPy~\autocite{2020SciPy-NMeth} library. The null hypothesis for this test assumes that the method rankings are random.
To enable a fair, isolated comparison, we use the same mid-sagittal plane estimate (by \textit{mri\_cc}) for all methods.

\begin{table*}[ht]
    \centering
    \rowcolors{6}{white}{lightgray}
    \begin{tabular}{|l|cc|cc|cc|cc|}
    \hline
    \textbf{Method} 
        & \multicolumn{4}{c|}{\textbf{Test set random ($N=30$)}} 
        & \multicolumn{4}{c|}{\textbf{Test set challenging ($N=19$)}} \\
    \cline{2-9}
    & \multicolumn{2}{c|}{\textbf{AC}} & \multicolumn{2}{c|}{\textbf{PC}} 
    & \multicolumn{2}{c|}{\textbf{AC}} & \multicolumn{2}{c|}{\textbf{PC}} \\
    \cline{2-9}
    & Mean & Median & Mean & Median & Mean & Median & Mean & Median \\
    \hline
    \textbf{FastSurfer-CC} 
        & \textbf{1.10} & 0.90 & \textbf{0.91} & \textbf{0.65} & \textbf{1.35} & 0.97 & \textbf{0.92} & \textbf{0.60} \\
    \textbf{acpc\_detect} 
        & 4.77 & \textbf{0.71} & 5.54 & \textbf{0.65} & 8.84 & \textbf{0.71} & 8.51 & 0.76 \\
    \textbf{SyN (ANTs)} 
        & 2.49 & 1.84 & 2.08 & 1.70 & 3.23 & 2.46 & 2.84 & 1.94 \\
    
    \hline
    \end{tabular}
    \vspace{0.5em}
    \caption{Mean and median localization error for anterior \& posterior commissure on the random and challenging test sets.}
    \label{tab:localization_acpc}
\end{table*}

\subsubsection{Comparison of group difference for corpus callosum thickness}

We perform a whole pipeline comparison, where we aim to find group differences in corpus callosum thickness between FastSurfer-CC and the only other method capable of generating thickness profiles: CCSeg. Here the combination of mid-sagittal plane selection, segmentation, and thickness analysis all contribute to the accurate final analysis. \R{R:failures}\defquotedtext{Q:failures}{After running both methods, we discard all cases where either method fails (did not produce outputs). Overall we discard 2.5\% of cases processed with CCSeg and no cases for FastSurfer-CC}. For each of the 100 thickness values and for each method, we create a linear model that predicts thickness with age, sex, and total brain volume (from FastSurfer) as covariates. \R{R:methods_multiple_comparison}The resulting p-values are then corrected with the Benjamini and Hochberg procedure for multiple comparisons~\autocite{benjamini1995controlling} and mapped onto the level sets of a template corpus callosum contour. The CC shape is colored according to the values of the level sets, and values between level sets are interpolated. 
Finally, we use the linear modeling with covariates from above to create separate models for each of the CC measures derived by our method and investigate which measures best explain group differences between patients and controls.
We use the SciPy~\autocite{2020SciPy-NMeth} and statsmodels~\autocite{seabold2010statsmodels} libraries for statistical testing.

\subsubsection{Reliability of corpus callosum thickness estimates and shape metrics}
\R{R:ICC_methods}\defquotedtext{Q:ICC_methods}{
Finally, we assess the reliability of our method by examining whether it produces consistent morphometric measures when applied to two scans of the same participant (test-retest analysis). We also compare our method with CCSeg. Because true morphological changes between scans are expected to be minimal, any observed differences are assumed to reflect inconsistencies in the analysis and acquisition. We quantify these differences using the intraclass correlation coefficient (ICC, ~\cite{mcgraw1996forming}) for all 100 thickness estimates, as well as for the summary metrics derived by our method. The ICC measures the similarity of the thickness values along the corpus callosum and the consistency of the shape metrics, with a value of 1 indicating perfect reproducibility. We specifically use the ICC for absolute agreement (criterion-referenced reliability), which assesses equality between measurements rather than mere correlation.
}
\ifmulticol
\\\
\\\
\\\
\\\
\fi

\section{Results}

First, we evaluate methods for mid-sagittal plane positioning. Then, we determine the accuracy of our trained AC \& PC localization and CC \& FN segmentation models and benchmark them against state-of-the-art approaches using manually annotated cases. Finally, we evaluate the full framework on the PREDICT-HD dataset, where we use all derived down-stream measures for statistical analysis.

\subsection{Mid-sagittal plane positioning}
\label{sec:midplane_results}

\R{R:correction4}\defquotedtext{Q:correction4}{We compare five methods for mid-sagittal plane positioning: i) finding the left-right symmetry axis by mid-space registration of FreeSurfer’s \textit{mri\_robust\_register}~\autocite{reuter2010highly} to a left-right flipped version of the same volume, ii) FreeSurfer’s \textit{mri\_cc}, iii) Yuki, iv) CCSeg, to v) FastSurfer-CC (our method).} \R{R:plane_comparison_explanation}\defquotedtext{Q:plane_comparison_explanation}{We present the results in Table~\ref{tab:midplane_rating}, which shows the number of times a method was chosen to provide a superior mid-sagittal plane for CC analysis (when no decision could be made, no vote was cast).} FastSurfer-CC outperforms all methods except for \textit{mri\_cc}. Here, method rankings disagree between raters. However, FastSurfer-CC is more than an order of magnitude faster than \textit{mri\_cc}.

\begin{figure*}[ht]
    \centering
    \ifmulticol
    \includegraphics[width=0.95\linewidth]{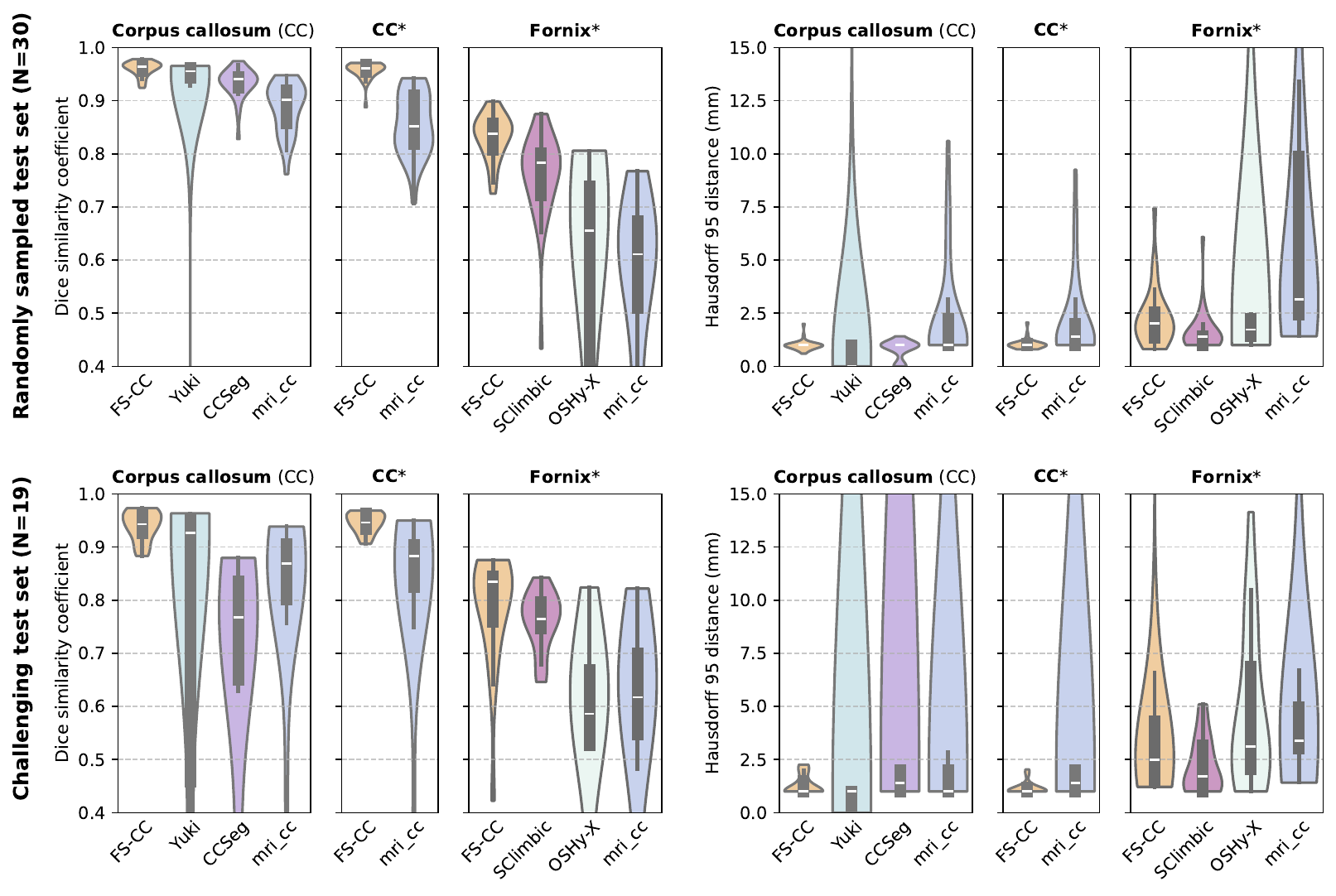}
    \else
    \includegraphics[width=1\linewidth]{figures/dice_hd_comparison_violinplots.pdf}
    \fi
    \caption{Dice similarity coefficient (DSC) and Hausdorff distance (HD95) for different evaluation scenarios. Shown are violin plots (colored), with additional boxplots in the center (dark grey). All differences in DSC  to our method are significant ($p<0.005$). A star (*) marks volumetric comparison at 5 mm width around the midslice; otherwise, only the midslice is considered. FS-CC: FastSurfer-CC}
    \label{fig:dice}
\end{figure*}

\subsection{Anterior- and posterior commissure localization}

We compare the localization accuracy of three methods (Table \ref{tab:localization_acpc}): i) acpc\_detect (part of the ART toolbox and the Yuki tool; \cite{ardekani2009model}), ii) non-linear multi-template registration using SyN (proposed and validated by~\cite{pallavaram2009validation}, \cite{liu2014automatic}, part of the ANTs toolbox;~\cite{avants2008symmetric,tustison2021antsx}), iii) FastSurfer-CC (ours). We observe that acpc\_detect is often highly accurate (\textless\SI{1}{mm} median error), but it produces 5 cases (10\%) with extreme failures (\textgreater\SI{10}{mm} error, for both AC \& PC). %
Errors of this magnitude would prevent further processing for many tasks and may introduce processing bias. Registration with SyN does not show failure cases of this magnitude but generally under-performs our method. As expected, localization methods perform much worse on the challenging test-set. Our method is a notable exception to that rule. Overall, we conclude that our method outperforms the other approaches, especially with respect to method robustness.

\subsection{Segmentation}

We compare four segmentation methods for CC and FN each. For the CC, we compare i) \mbox{FastSurfer-CC} (Ours), ii) CCSeg, iii) Yuki, and iv) \textit{mri\_cc}. For the FN, we compare i) \mbox{FastSurfer-CC} (Ours), ii) SClimbic, iii) \textit{mri\_cc}, and iv) OSHy-X. For all evaluations, we present the Dice similarity coefficient (DSC) and 95th percentile Hausdorff distance (HD95) with respect to manual segmentation in Figure~\ref{fig:dice}. The CC segmentation analysis is further sub-divided into 1.\ area (only the mid-sagittal plane) and 2.\ volume (only the 5 mid-sagittal slices segmented by \textit{mri\_cc}). Note that only FastSurfer-CC and FreeSurfer's \textit{mri\_cc} support multi-slice evaluations.

FastSurfer-CC significantly outperforms ($p<0.01$) other methods in CC segmentation DSC for both the randomly selected and challenging test scenarios. FastSurfer-CC accuracy also only decreases slightly when using more challenging data -- especially compared to other methods.
While significantly worse than FastSurfer-CC in DSC, CCSeg narrowly outperforms FastSurfer-CC in Hausdorff distances on the random test-set (mean \SI{0.77}{mm} and \SI{0.9}{mm}, $p<0.05$). However, on the challenging test-set CCSeg performs worst of all methods with a high Hausdorff Distance driven by outliers (mean \SI{13.43}{mm} and \SI{1.29}{mm}), while our method outperforms all others, retaining similar accuracy as seen on the randomly selected cases ($p\approx0.05$).

For FN segmentation, FastSurfer-CC also achieves significantly higher DSC than other methods for both randomly selected and challenging test scenarios ($p<0.01$). \R{R:correction5}\defquotedtext{Q:correction5}{Similar to CC segmentation, the FN segmentation is remarkably reliable even in challenging scenarios with average $\text{DSC}>0.8$.}
While  FastSurfer-CC outperforms SClimbic (specialized to fornix) on both test-set and difficult test-set for DSC, the situation is reversed for Hausdorff distances (mean \SI{1.63}{mm} and \SI{2.29}{mm}). Overall, we find that our method delivers superior CC segmentations and is also more robust than other methods.

\begin{figure}[ht]
    \centering
    \ifmulticol
    \includegraphics[width=1\linewidth, trim={7.5cm 0.7cm 0.8cm 0},clip]{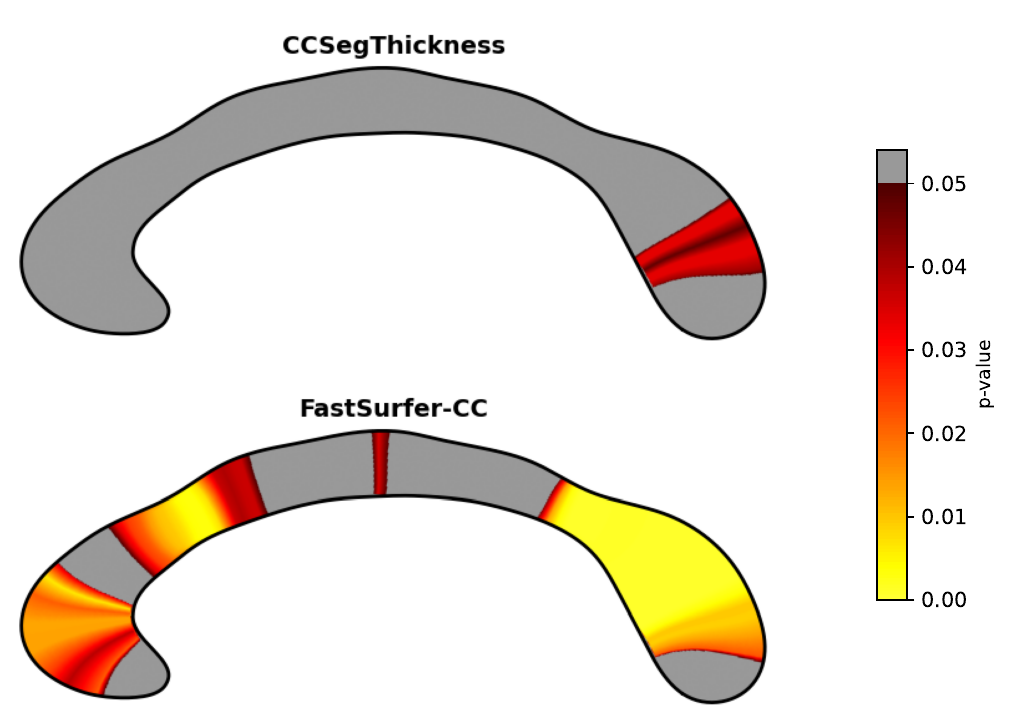}
    \else
    \includegraphics[width=0.6\linewidth, trim={7.5cm 0.7cm 0.8cm 0},clip]{figures/cc_comparison_new.pdf}
    \fi
    \caption{Corrected p-values for the group comparison of corpus callosum thickness between Huntington's disease patients and controls mapped onto a template.}
    \label{fig:thickness_comparison}
\end{figure}

\begin{figure}[ht]
    \centering
    \ifmulticol
    \includegraphics[width=0.95\linewidth]{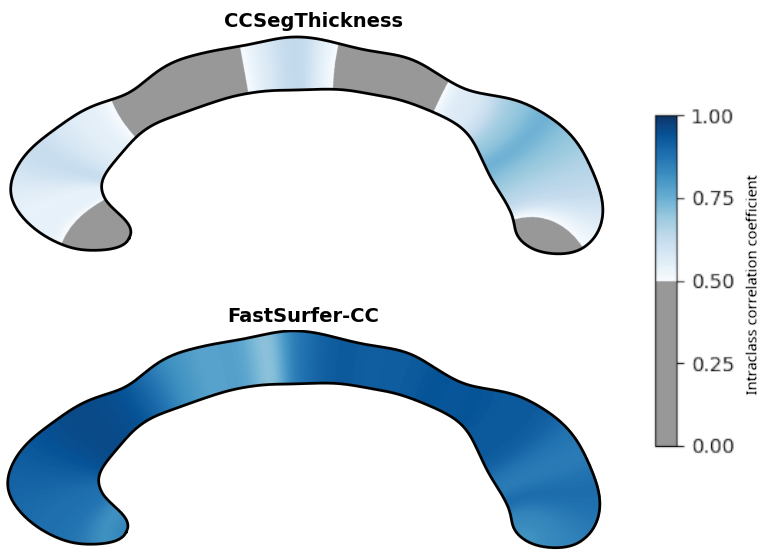}
    \else
    \includegraphics[width=0.6\linewidth]{figures/ICC.png}
    \fi
    
    \caption{Intraclass correlation coefficient for a test-retest analysis in the PREDICT-HD dataset. Higher scores (darker blue) indicate better reproducibility of thickness values within a scan sessions.}
    \label{fig:icc_thickness}
\end{figure}

\begin{figure*}
    \centering\includegraphics[width=1\linewidth]{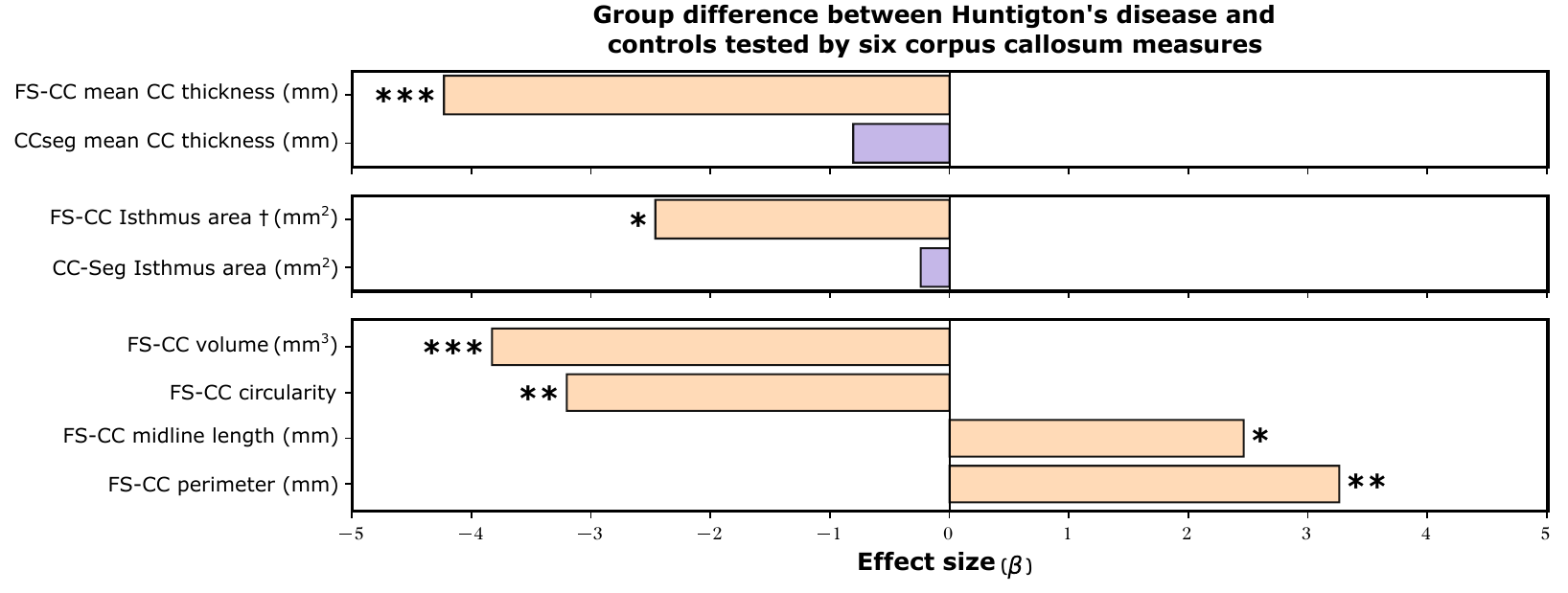}
    \caption{Effect sizes ($\beta$ values in the linear model) and p-values for a comparison of Huntington's disease patients with controls. Negative values mean reduction of measure in patients. $\dagger$ $\rightarrow$ We use FastSurfer-CC's improved sub-division method. \mbox{* $\rightarrow p<0.05$}, ** $\rightarrow p<0.01$, *** $\rightarrow p<0.001$, FS-CC $\rightarrow$ FastSurfer-CC}
    \label{fig:effect_sizes}
\end{figure*}

\subsection{Corpus callosum morphometry in Huntington's disease}
\label{sec:HD_analysis}

To validate that our method can find group differences, we perform the previously introduced statistical analysis, comparing Huntington's disease patients with controls. We use CCSeg and FastSurfer-CC methods end-to-end to generate thickness profiles (as they are the only methods with this capability) and other corpus callosum measures. We show a plot with corrected p-values in Figure~\ref{fig:thickness_comparison}. While only 6 of 100 thickness values show significant effects for CCseg ($p<0.05$, dark red in Fig.~\ref{fig:thickness_comparison}), our method finds significant effects in 57 of 100 values.%

The Hofer-Frahm sub-segmentation yields no significant effects when using CCSeg for segmentation and post-processing ($p>0.05$ for each sub-segment). \R{R:grammar_fix_and_HD_interpretation}\defquotedtext{Q:grammar_fix_and_HD_interpretation}{The sub-segmentation by FastSurfer-CC, however, shows significant changes in the isthmus area ($p<0.05$, second posterior segment), aligning with changes in CC thickness (see Fig.~\ref{fig:thickness_comparison}) and previous work~\autocite{di2012multimodal}.} When using our shape-aware subdivision method instead of the standard Hofer-Frahm scheme (while maintaining sub-division ratios), the significance of this finding increases (from $p\approx0.04$ to $p\approx0.01$). 

Furthermore, our model provides an extensive library of measures of corpus callosum shape. The average thickness, for example, shows the expected highly significant difference using our method ($p<0.0001$), but not for CCSeg ($p\approx0.42$) using the same statistical model as above. Other measures only provided by our method also show significant differences: CC volume at 5 mm width ($p<0.001$), total CC perimeter ($p<0.01$), CC circularity ($p<0.01$), and the length of the intercallosal line ($p<0.05$). We find no significant differences for the total CC area on the mid-sagittal plane ($p\approx0.27$), the CC curvature (bend of the intercallosal line, $p\approx0.39$) and the CC index ($p\approx0.87$). We show an overview of these effects in Figure~\ref{fig:effect_sizes}. \R{R:HD_interpretation}\defquotedtext{Q:HD_interpretation}{These findings align with the previously known~\autocite{rosas2010altered, di2012multimodal} atrophy of the corpus callosum during HD progression. Local changes in CC thickness, especially in the posterior CC section were also observed by~\cite{rosas2010altered, di2012multimodal} and can serve as an early indicator of white matter demyelination damage in pre-clinical HD.}

\subsection{Reliability of corpus callosum morphometry}
\label{sec:ICC}
\R{R:ICC_results}\defquotedtext{Q:ICC_results}{In Figure~\ref{fig:icc_thickness}, we show the test-retest reliability of CC thickness values of our method and CCSeg (high ICC values indicate high reliability). Our method shows high test-retest reliability across all CC sections, while CCSeg demonstrates very low ICC especially on less curved sections and CC endpoints. In addition to the evaluation of CC thickness estimates, we provide an overview of the reliability of the corpus callosum metrics in the Appendix Figure~\ref{fig:metrics_icc}, where all metrics derived with our method show high reliability ($\text{ICC}>0.7$).}

\section{Discussion}

Overall, we propose a fast, robust, and accurate framework for corpus callosum morphometry. We have validated its components individually and also showed that it outperforms the state-of-the-art with respect to sensitivity for a group comparison of Huntington's disease.
Furthermore, our method offers improved sub-segmentation approaches and additional downstream metrics. 
Besides corpus callosum analysis, our framework can be used for head pose normalization, which provides a fast alternative to rigid template registration and can aid as initialization of other affine or non-linear registration tasks. Since the alignment takes only seconds, it can be employed to preprocess large datasets and is applicable in real-time applications, e.g., for quality control of orientation angles in diffusion MRI.

Our evaluation is specifically designed to consider the performance on unseen data of the difficult test-set to evaluate critical method robustness. This includes cases with brain lesions, strong atrophy, and artifacts, which were unseen during training.  
Our experiments and results demonstrate that both our method and previous state-of-the-art methods show excellent average performance on randomly sampled cases. In a preliminary setting, we established an inter-rater reliability for segmentation of 0.950 (CC DSC) and 0.852 (FN DSC). With performance around that same level for our method, further distinction between methods is increasingly hampered by this inter-rater reliability threshold. Equally, the localization is highly accurate (typically less than one voxel error). The performance of the state-of-the-art methods Yuki, CCSeg, and acpc\_detect is similar, yet slightly lower on high-quality data.\R{R:no_DL} \defquotedtext{Q:no_DL}{We note that our method is the only deep learning–based segmentation approach included in the comparison, reflecting the limited availability of accessible and reusable deep learning methods for corpus callosum analysis and its related sub-tasks. Consequently, some of the observed improvements are likely attributable to the adoption of this more recent methodological paradigm.}

Nonetheless, our analysis shows that competing methods regularly exhibit extreme errors -- both for challenging and randomly selected cases. While large errors can often be caught in quality control, more subtle errors may be missed and can bias down-stream analysis. Therefore, we highlight that FastSurfer-CC's accuracy remains high across the board when moving from randomly selected to challenging cases while at the same time outperforming specialized methods in good conditions.\R{R:limitations} \defquotedtext{Q:limitations}{One limitation of the reference-based method comparisons (segmentation and localization accuracy) is the potential for rater bias. Because some raters annotated cases across the training, validation, and test sets, the model may have learned rater-specific annotation preferences, potentially resulting in overly optimistic performance estimates. However, given the relatively simple task of segmenting the corpus callosum, which typically has a clearly defined boundary, the impact of this bias is expected to be small.} 

\R{R:HD_discussion}\defquotedtext{Q:HD_discussion}{Furthermore, beyond the refrence-based method comparisons, however, our method also demonstrates increased sensitivity in a cohort of Huntington's disease patients and controls, consistent with previously known findings~\autocite{rosas2010altered, di2012multimodal}. Our method shows the expected atrophy in the CC isthmus and general atrophy measured by reduced thickness, volume at 5 mm width, perimeter, and length. We also observe that the theoretical improvements to the sub-segmentation method translate to higher sensitivity in this analysis. While many of the proposed summary metrics are markers for general atrophy, they are not necessarily highly correlated (see Appendix~\ref{appendix:metrics_corr}) indicating an opportunity for future research to identify more specific markers for CC degeneration. CC circularity, for example, is a combined metric (area divided by squared boundary length), which has been established as an early marker for Alzheimer's disease~\autocite{van2018callosal}.} \R{R:multiple_comparisons}\defquotedtext{Q:multiple_comparisons}{Besides these summary metrics, FastSurfer-CC also provides local thickness estimates. In the comparison of Huntington's disease patients and controls, local CC thickness differences demonstrate a higher sensitivity than differences based on CC sub-areas (see~Section~\ref{sec:HD_analysis}) -- even when including 100 CC thickness estimates and adjusting for multiple comparisons~\autocite{benjamini1995controlling}.}
\R{R:discussion_trt}\defquotedtext{Q:discussion_trt}{Finally, our method also demonstrates high test-rest reliability in healthy and diseased cases of the PREDICT-HD cohort, showing highly consistent corpus callosum segmentation and thickness estimation within scan sessions}.

Our contributions provide researchers with a highly sensitive, reliable, and novel tool to explore and analyze CC shape changes in disease and aging. While our framework proposes new, sensitive markers of length and curvature of the CC as well as a novel geometry-aware sub-segmentation scheme, it also provides easy access to a plethora of previously proposed metrics and schemes.
FastSurfer-CC is part of the open-source project FastSurfer at \url{github.com/DeepMI/FastSurfer} and enhances FastSurfer with rapid head pose standardization, AC \& PC localization, CC \& FN segmentation, CC thickness profiles, and the other new morphometrics. With less than 10 seconds processing time, this contribution will also improve FastSurfer's overall processing speed.

\section{Acknowledgments}

We are grateful to Dr.\ Jingjing Wu for her advice and support on neuroanatomy topics. We also thank Andreas Girodi for his dedication and support in the initial stages of the project.
Furthermore, we thank the Rhineland Study group (Monique M.B.\ Breteler), the group of Toni St\"ocker, and the group of Theodor R\"uber  for supporting the data acquisition and management. 

This work is supported by DZNE institutional funds, the Federal Ministry of Research, Technology and Space of Germany (BMFTR 031L0206), the NIH (R01-AG064027, R01 MH131586, R01 MH130899), the Chan-Zuckerberg Initiatives Essential Open Source Software for Science RFA (EOSS5 2022-252594), the Helmholtz Foundation Model Initiative (The Human Radiome Project), and the Ministry of Culture and Science North Rhine-Westphalia's Initiative InVirtuo 4.0 (PB22-063A).

Data used in this work was generously provided by the participants in PREDICT-HD and made available by the PREDICT-HD Investigators and Coordinators of the Huntington Study Group, Jane Paulsen, Principal Investigator. PREDICT-HD was funded by the National Institute of Health (NIH) under Grant\# NS040068. Data used in the preparation of this article were obtained in part by the OASIS Cross-Sectional with principal investigators D.\ Marcus, R. Buckner, J.\ Csernansky, J.\ Morris; P50 AG05681, P01 AG03991, P01 AG026276, R01 AG021910, P20 MH071616, U24 RR021382, and OASIS: Longitudinal: Principal Investigators: D.\ Marcus, R.\ Buckner, J.\ Csernansky, J.\ Morris; P50 AG05681, P01 AG03991, P01 AG026276, R01 AG021910, P20 MH071616, U24 RR021382. Further, data used in the preparation of this article were obtained from the MIRIAD database. The MIRIAD investigators did not participate in analysis or writing of this report. The MIRIAD dataset is made available through the support of the UK Alzheimer’s Society (Grant RF116). The original data collection was funded through an unrestricted educational grant from GlaxoSmithKline (Grant 6GKC). Data used in preparation of this article were obtained from the Alzheimer’s Disease Neuroimaging Initiative (ADNI) database (\url{adni.loni.usc.edu}). As such, the investigators within the ADNI contributed to the design and implementation of ADNI and/or provided data but did not participate in analysis or writing of this report. A complete listing of ADNI investigators can be found at: \url{http://adni.loni.usc.edu/wp-content/uploads/how_to_apply/ADNI_Acknowledgement_List.pdf}. Data collection and sharing for this project was funded by the Alzheimer’s Disease Neuroimaging Initiative (ADNI) (National Institutes of Health Grant U01 AG024904) and DOD ADNI (Department of Defense award number W81XWH-12-2-0012). ADNI is funded by the National Institute on Aging, the National Institute of Biomedical Imaging and Bioengineering, and through generous contributions from the following: AbbVie, Alzheimer’s Association; Alzheimer’s Drug Discovery Foundation; Araclon Biotech; BioClinica, Inc.; Biogen; Bristol-Myers Squibb Company; CereSpir, Inc.; Cogstate; Eisai Inc.; Elan Pharmaceuticals, Inc.; Eli Lilly and Company; EuroImmun; F. Hoffmann-La Roche Ltd and its affiliated company Genentech, Inc.; Fujirebio; GE Healthcare; IXICO Ltd.; Janssen Alzheimer Immunotherapy Research \& Development, LLC.; Johnson \& Johnson Pharmaceutical Research \& Development LLC.; Lumosity; Lundbeck; Merck \& Co., Inc.; Meso Scale Diagnostics, LLC.; NeuroRx Research; Neurotrack Technologies; Novartis Pharmaceuticals Corporation; Pfizer Inc.; Piramal Imaging; Servier; Takeda Pharmaceutical Company; and Transition Therapeutics. The Canadian Institutes of Health Research is providing funds to support ADNI clinical sites in Canada. Private sector contributions are facilitated by the Foundation for the National Institutes of Health (\url{www.fnih.org}). The grantee organization is the Northern California Institute for Research and Education, and the study is coordinated by the Alzheimer’s Therapeutic Research Institute at the University of Southern California. ADNI data are disseminated by the Laboratory for Neuro Imaging at the University of Southern California. Data were also provided in part by the Human Connectome Project, WU-Minn Consortium (Principal Investigators: David Van Essen and Kamil Ugurbil; 1U54MH091657) funded by the 16 NIH Institutes and Centers that support the NIH Blueprint for Neuroscience Research; and by the McDonnell Center for Systems Neuroscience at Washington University.

\section{Data \& code availability Statement}

The source code for FastSurfer-CC is part of the FastSurfer toolbox at \url{https://github.com/Deep-MI/FastSurfer}. The rating tool used for the user study is based on~\url{https://github.com/deep-MI/segmentation_labeling}.

All MRI datasets, except for data of the Rhineland Study and in-house datasets, are publicly available, and references to the open-source repositories are provided in Appendix Table~\ref{tab:split}. The UK-Bonn dataset is not publicly available since it contains information that could compromise the privacy of research participants. Access to the 7T dataset can be provided to scientists upon reasonable request to the Human MRI department of the German Center for Neurodegenerative Diseases. Data from the Rhineland Study is not publicly available because of data protection regulations; however, access can be provided to scientists in accordance with the Rhineland Study’s Data Use and Access Policy. Requests to access the data should be directed to Dr.\ Monique Breteler at \url{RS-DUAC@dzne.de}.

\section{Author contributions}
\noindent
\textbf{Clemens Pollak}: Methodology, Conceptualization, Formal analysis, Investigation, Data Curation, Software, Validation, Writing – original draft, Writing – review \& editing, Visualization.\\
\textbf{Kersten Diers}: Conceptualization, Data Curation, Resources, Writing – review \& editing. \\
\textbf{Santiago Estrada}: Data Curation, Resources, Writing – review \& editing, Visualization. \\
\textbf{David K\"ugler}: Conceptualization, Writing – original draft, Writing – review \& editing, Supervision \\
\textbf{Martin Reuter}: Conceptualization, Methodology, Software, Resources, Writing – original draft, Writing – review \& editing, Supervision, Project administration, Funding acquisition.

\section{Declaration of competing interests}
\noindent
The authors do not declare any competing interests.

\vfill
\pagebreak

\appendix
\renewcommand{\thefigure}{A.\arabic{figure}}
\renewcommand{\thetable}{A.\arabic{table}}
\setcounter{figure}{0}

\section{Appendix}

\subsection{Data description}

For method development, we manually labeled 177 cases with corpus callosum, fornix, anterior and posterior commissures, and split them into training, validation, and test-set. We also established another difficult test-set with cases that are especially challenging due to brain lesions, motion artifacts, strong atrophy, and low imaging contrast. For labeling, we use mid-sagittal planes generated by FreeSurfer's \textit{mri\_cc} tool. Since this FreeSurfer tool uses whole brain segmentations, we quality control the whole brain segmentations prior to processing and use lesion inpainting \autocite{pollak2025fastsurfer} when large areas of damaged tissues are present to circumvent errors. We show an overview of the datasets used for annotation in Table~\ref{tab:split}.

\begin{table*}[ht]
\centering
\begin{tabular}{|l|c|c|c|c|c|}
\hline
\textbf{Dataset} & \textbf{\# Train} & \textbf{\# Val} & \textbf{\# Test} & \textbf{\# Diffic.} & \textbf{\# Total} \\ 
\hline
ABIDE \autocite{di2014autism}           & 7  & 2 & 4 & 0 & 14 \\
ABIDE-II~\autocite{di2017enhancing}    & 3 & 1 & 0 & 0 & 4 \\
ADNI \autocite{jack2008adni}            & 15 & 7 & 3 & 6 & 32  \\
LA5C \autocite{poldrack2016phenome}            & 9 & 5 & 0 & 2 & 17\\
MIRIAD \autocite{malone2013miriad}          & 10 & 2 & 3 & 1 & 17 \\
OASIS1 \autocite{marcus2007open}            & 8 & 4 & 5 & 1 & 18 \\
OASIS2 \autocite{marcus2010open}            & 8 & 2 & 2 & 2 & 14 \\
Rhineland Study (\SI{1}{\milli\meter}, \cite{koch2023versatile})             & 12 & 2 & 4 & 0 & 20 \\
Rhineland Study \autocite{koch2023versatile}          & 9 & 3 & 7 & 1 & 20 \\
HCP \autocite{van2012human, bookheimer2019lifespan}             & 5 & 3 & 0 & 1 & 9 \\
IXI \autocite{IXI} & 0 & 0 & 0 & 2 & 2 \\
7T (in-house)             & 7 & 0 & 2 & 0 & 9 \\
UK-Bonn epilepsy (in-house)             & 0 & 0 & 0 & 3 & 3 \\ \hline
Combined               & 93 & 31 & 30 & 19 & 173 \\ \hline
\end{tabular}
\caption{Overview of the used datasets and their inclusion in the sub-sets: training, validation, test, and difficult test.}
\label{tab:split}
\end{table*}

\begin{table*}[ht]
\centering
\setlength{\tabcolsep}{3.7pt} 
\begin{tabular}{|l|cccccccc|ccc|cccc|ccc|ccc|}
\multicolumn{1}{c}{} & \multicolumn{8}{c}{\textbf{Diagnosis}} & \multicolumn{3}{c}{\textbf{Sex}} & \multicolumn{4}{c}{\textbf{Age}} & \multicolumn{3}{c}{\textbf{Field (T)}} & \multicolumn{3}{c}{\textbf{Manufacturer}} \\ 
\hline
 & AD & AU & C & DE & EP & MC & SC & U & F & M & U & $\mathbf{\mu}$ & $\mathbf{\sigma}$ & min & max & 1.5 & 3.0 & 7.0 & GE & Phil & Siem \\ 
\hline
\textbf{Train}    & 0 & 5 & 32 & 17 & 0 & 1 & 5 & 33 & 46 & 43 & 4  & 52.2 & 21.3 & 18 & 90 & 26 & 60  & 7 & 11 & 7  & 75 \\ 
\textbf{Val.}     & 0 & 2 & 16 & 3  & 0 & 0 & 2 & 8  & 20 & 10 & 1  & 50.3 & 22.0 & 20 & 81 & 8  & 23  & 0 & 2  & 3  & 26 \\ 
\textbf{Test}     & 0 & 2 & 13 & 2  & 0 & 0 & 0 & 13 & 13 & 15 & 2  & 53.1 & 22.0 & 19 & 88 & 10 & 18  & 2 & 4  & 2  & 24 \\ 
\textbf{Diffi.}   & 1 & 0 & 3  & 6  & 3 & 1 & 1 & 4  & 7  & 8  & 4  & 64.0 & 24.0 & 22 & 88 & 6  & 13  & 0 & 1  & 2  & 16 \\ 
\hline
\textbf{Combi.} & 1 & 9 & 64 & 28 & 3 & 2 & 8 & 58 & 86 & 76 & 11 & 53.1 & 21.1 & 18 & 90 & 50 & 114 & 9 & 18 & 14 & 141 \\
\hline
\end{tabular}
\caption{Overview of the dataset used for method development and evaluation. AD: ADHD (Attention-Deficit/Hyperactivity Disorder), AU: Autism, C: Control, DE: Dementia, EP: Epilepsy, MC: MCI (Mild Cognitive Impairment), SC: Schizophrenia, U: Unknown, F: Female, M: Male, $\mathbf{\mu}$: Mean (Age), $\mathbf{\sigma}$: Standard Deviation (Age), Field: Magnetic field strength of MRI scanner, GE: General Electric, Phil: Philips, Siem: Siemens}
\label{tab:dev_sets_metadata}
\end{table*}

\begin{table*}[ht]
\centering
\setlength{\tabcolsep}{3.7pt} 
\begin{tabular}{|l|ccc|cccc|cc|ccc|}
\multicolumn{1}{c}{} & \multicolumn{3}{c}{\textbf{Sex}} & \multicolumn{4}{c}{\textbf{Age}} & \multicolumn{2}{c}{\textbf{Field (T)}} & \multicolumn{3}{c}{\textbf{Manufacturer}} \\ 
\hline
 & F & M & U & $\mathbf{\mu}$ & $\mathbf{\sigma}$ & min & max & 1.5 & 3.0 & GE & Phil & Siem \\ 
\hline
\textbf{HD}      & 634 & 358 & 0 & 42.1 & 11.1 & 19 & 78 & 623 & 369 & 528 & 93 & 369 \\ 
\textbf{Control} & 184 & 92  & 0 & 46.1 & 12.0 & 19 & 85 & 154 & 122 & 133 & 25 & 118 \\ 
\hline
\textbf{Combined}& 818 & 450 & 0 & 43.0 & 11.5 & 19 & 85 & 777 & 491 & 661 & 118 & 487 \\
\hline
\end{tabular}
\caption{Overview of the dataset used for group comparisons of Huntington's disease cases and controls. The row marked with HD lists participants that tested positive for the gene expansion associated with Huntington's disease and have a CAG (cytosine, adenine, and guanine) repeat of 36 or greater. F: Female, M: Male, U: Unknown, $\mathbf{\mu}$: Mean (Age), $\mathbf{\sigma}$: Standard Deviation (Age), Field: Magnetic field strength of MRI scanner, GE: General Electric, Phil: Philips, Siem: Siemens}
\label{tab:appendix_predicthd}
\end{table*}

\subsection{Calculating corpus callosum thickness, length and curvature}

\label{appendix:laplace}

To derive accurate morphometrics from the CC segmentation, we first detach the corpus callosum contour from the voxel grid by applying a Gaussian filter to the binary label. Next, we apply the marching squares algorithm implemented in the scikit-image library~\autocite{scikit-image}. Next, the closed boundary contour is triangulated with the meshpy library~\autocite{MeshPyWebsite}, providing a high-quality, dense triangle mesh of the CC shape (see Fig.~\ref{fig:levelsets}). 
To split the CC boundary into an inferior and superior curve at its end-points, we use the relative position of the anterior and posterior commissures. In contrast to previous heuristics (e.g., using the curvature of the contour, \cite{adamson2011thickness}), the AC \& PC are better references because they are deformed together with the CC, leading to better intercallosal lines in brains with strong atrophy or other disease-related changes. 
Based on the training set, we design a heuristic that gives visually accurate endpoints by first finding intermediate anchor points with respect to the AC-PC line %
and then selecting the closest points on the contour as anterior and posterior endpoints.
After deriving the end-points, we split the contour into an upper and lower portion, apply appropriate boundary conditions (of $f_\text{inf/sup}=-1 / 1$ along the inferior and superior boundary and $f_\text{ep}=0$ at the two endpoints), and solve the Laplace equation $\Delta f = 0$ for all interior points on the mesh. We extend the LaPy Python library\footnote{https://github.com/deep-mi/LaPy}~\autocite{reuter2006laplace, wachinger2015brainprint} for these and the following operations. We extract the zero-level set of the Laplace solution as a piecewise linear path to obtain an \textit{intercallosal line}, running from one endpoint to the other. 
We then resample the intercallosal line to have exactly $N+2$ equidistant points ($N$ being the number of the desired interior thickness estimates, default $N=100$). %
Finally, to avoid tracing gradients for thickness estimates, we rotate the Laplace solution on the mesh so that the new level sets are orthogonal to the level sets of the original solution. This is done by rotating the gradients around the normals by 90 degrees, then solving the Poisson equation $ \Delta f = h$ with the divergence of the rotated gradient as $h$. It is now easy to extract the level set curves at the $N$ locations along the intercallosal line. The individual lengths of these level set curves provide the thickness profile of the CC. Furthermore, total CC length and curvature are computed directly from the intercallosal line and are also provided.

\subsection{Additional corpus callosum metrics}
\label{sec:additional_metrics}

In addition to thickness profiles, curvature, midline length, and sub-segmentation areas, we also include two previously proposed metrics for corpus callosum analysis and describe them here for documentation purposes.

\subsubsection{Circularity}

Circularity is a low-descriptive shape measure comparing area $A$ to boundary length $L$, which is maximized by the disc, where it becomes one:
\begin{equation}
   Circ = 4 \pi \frac{A}{L^2}.
\end{equation}
The corpus callosum circularity was proposed by \cite{van2018callosal} as a marker for Alzheimer's disease. CC atrophy and thinning from aging or disease will decrease area faster than boundary and therefore lead to a decreasing circularity. 
While our framework offers more fine-grained measures of length, thickness, and curvature individually, some researchers may desire an aggregated metric, which is why we include the corpus callosum circularity in the pipeline output.

\subsubsection{Corpus callosum index}

Similar to the corpus callosum circularity, the corpus callosum index is also an aggregate metric that tries to condense the complex shape changes of the CC by aggregating a few straightforward measurements that traditionally were easy to obtain manually. It was first proposed as a proxy for brain volume in the analysis of multiple sclerosis (MS) patients~\autocite{yaldizli2010corpus}. We have outlined the calculation in Figure~\ref{fig:cc_index}. Below are the detailed steps to calculate the index from a corpus callosum contour:

\begin{figure}[ht]
    \centering
    \ifmulticol
    \includegraphics[width=1\linewidth, trim={2.5cm 1.3cm 2.5cm 1.3cm},clip]{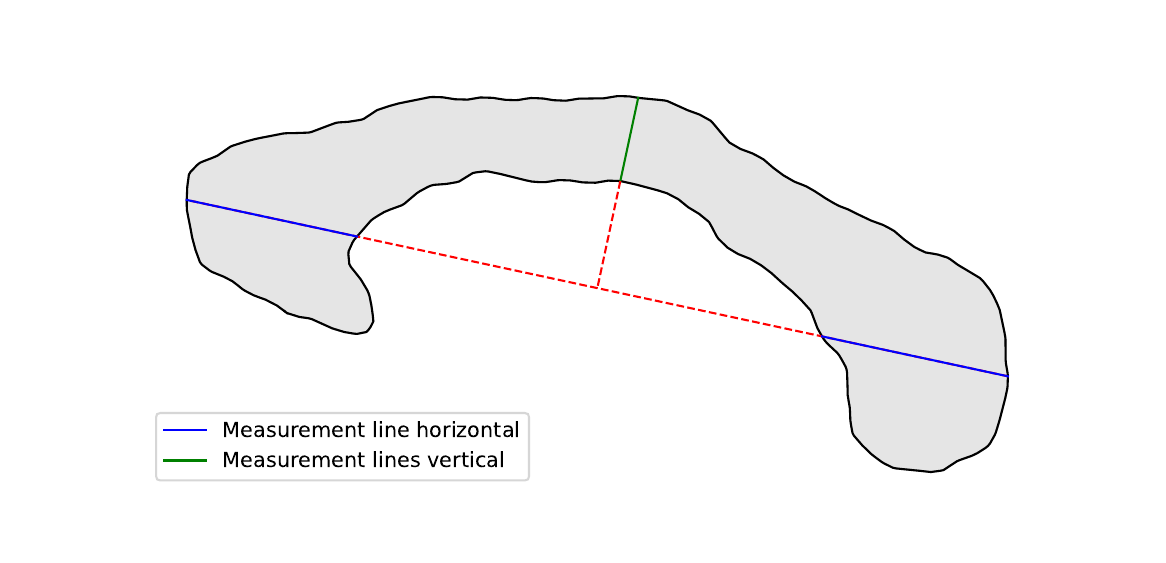}
    \else
    \includegraphics[width=0.6\linewidth, trim={2.5cm 1.3cm 2.5cm 1.3cm},clip]{figures/cc_index.pdf}
    \fi
    \caption{The corpus callosum index is defined as the sum of the two horizontal measurement segments (blue) and the vertical measurement segment (green).}
    \label{fig:cc_index}
\end{figure}

\begin{tight_enumerate}
    \item Draw the longest possible line between the anterior posterior edges of the CC (cutting the CC into 3 parts)
    \item Draw a second line perpendicular to the first and crossing the first line's midpoint (cutting the CC in the middle, into 4 parts overall).
    \item Measure the thickness of cuts at the three cutting locations and sum them.
\end{tight_enumerate}

\subsection{Metrics correlation}
\label{appendix:metrics_corr}

\begin{figure}[!ht]
    \centering
    \ifmulticol
    \includegraphics[width=1\linewidth]{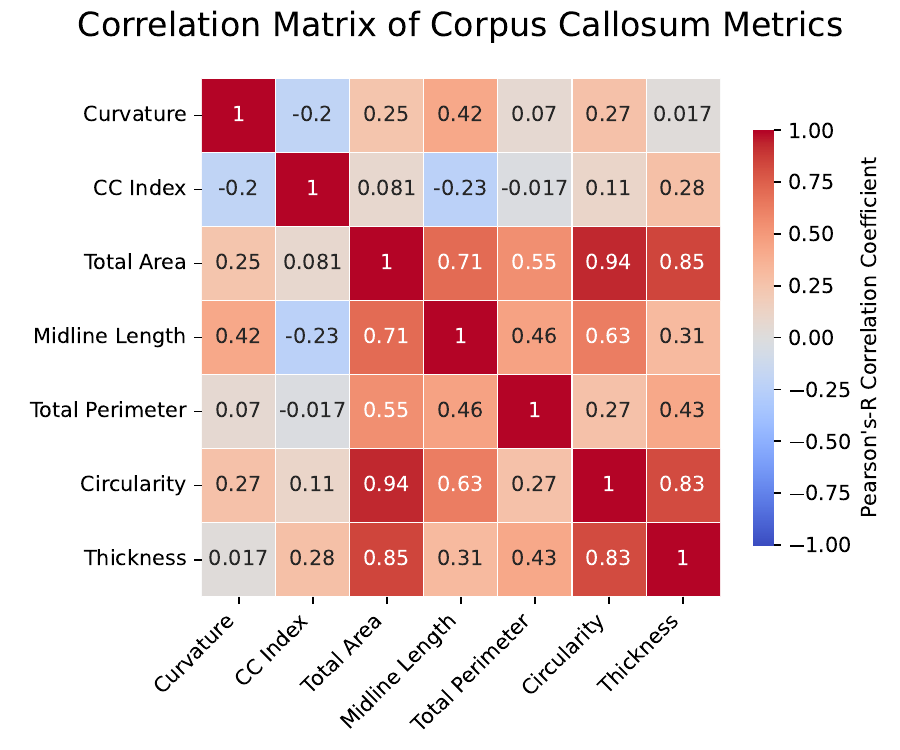}
    \else
    \includegraphics[width=0.6\linewidth]{figures/correlation_matrix.pdf}
    \fi
    \caption{Linear correlation coefficients (Pearson's-R) for corpus callosum metrics derived in the cohort of HD patients and controls.}
    \label{fig:corr_matrix}
\end{figure}

\R{R:metrics_correlation}\defquotedtext{Q:metrics_correlation}{
In Figure~\ref{fig:corr_matrix} we show the linear correlation coefficients (Pearsons's-R) for corpus callosum metrics derived in the cohort of HD patients and controls to evaluate whether some of the provided metrics are redundant. We observe that while some metrics are highly correlated (e.g.\ total area and circularity) many other metrics like curvature and CC index do not show high correlations across the board ($R<0.28$ and $R<0.42$ respectively). These low correlations indicate that the additional metrics are not necessarily redundant and could provide different information about CC morphometry. FastSurfer-CC offers multiple markers of corpus callosum anatomy, allowing researchers to select the most relevant subset for their specific questions.
}

\subsection{Reliability of corpus callosum metrics}
\label{appendix:metrics_icc}

\begin{figure}[!ht]
    \centering
    \ifmulticol
    \includegraphics[width=1\linewidth]{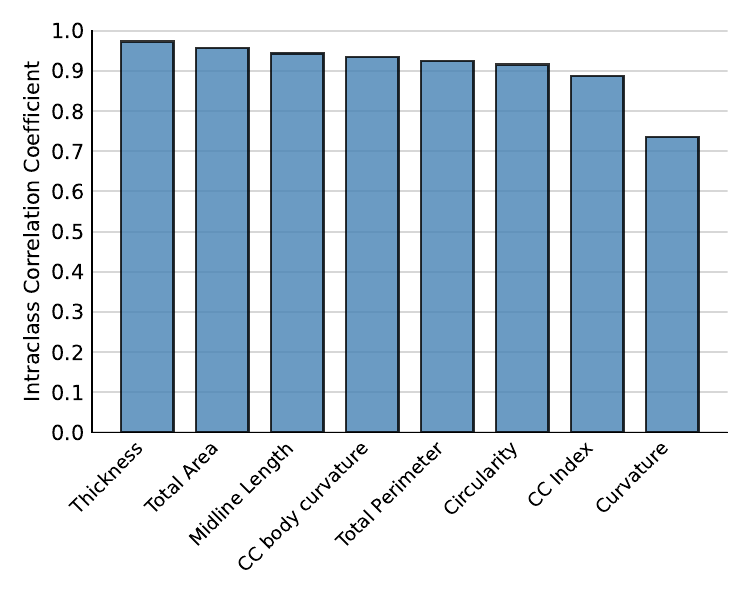}
    \else
    \includegraphics[width=0.6\linewidth]{figures/icc_values.pdf}
    \fi
    \caption{Intraclass correlation coefficients for corpus callosum metrics of FastSurfer-CC using two MRIs from the same scan session in the PREDICT-HD dataset.}
    \label{fig:metrics_icc}
\end{figure}

\R{R:metrics_icc}\defquotedtext{Q:metrics_icc}{
In addition to the test-retest analysis using thickness values (Section~\ref{sec:ICC}) we evaluate the test-retest reliability of the metrics derived with our method, using the same data and ICC metrics. In Figure~\ref{fig:metrics_icc} we show that metrics derived by our method are reliable across scans (ICC$>$0.85). The CC curvature (measured by the curvature of the midline) scores slightly lower on ICC (\textgreater0.7). We hypothesize, that this is driven by a potential inconsistency in the CC endpoints selection. For some anatomies the CC can lack a clear endpoint and present a flat shape at the end, especially in the posterior subsegment. To confirm this, we compute the curvature of the corpus callosum (CC) body by restricting the calculation to the central 70\% of the midline. This metric exhibits higher reliability, albeit at the cost of excluding curvature changes at the anterior and posterior ends of the CC.}

\subsection{Difficult examples}

\begin{figure*}[ht!] %
    \centering %

    \newlength{\imagewidth}
    \newlength{\rowindicatorwidth}
    \setlength{\rowindicatorwidth}{0.03\textwidth} %
    \setlength{\imagewidth}{0.315\textwidth} %

    \parbox[t]{\rowindicatorwidth}{}\hfill
    \parbox[t]{\imagewidth}{\centering \textbf{Original Image}}\hfill
    \parbox[t]{\imagewidth}{\centering \textbf{Rater Annotation}}\hfill
    \parbox[t]{\imagewidth}{\centering \textbf{Network Prediction}} \\
    \vspace{0.5em} %
    \hrule %
    \vspace{0.5em} %

    \parbox[t]{\rowindicatorwidth}{\centering \textbf{A}}\hfill
    \includegraphics[width=\imagewidth]{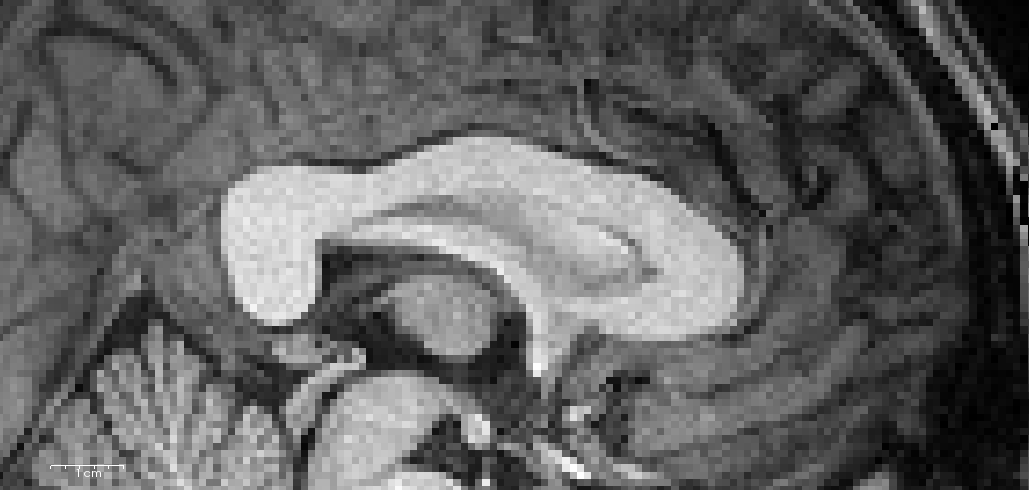}\hfill
    \includegraphics[width=\imagewidth]{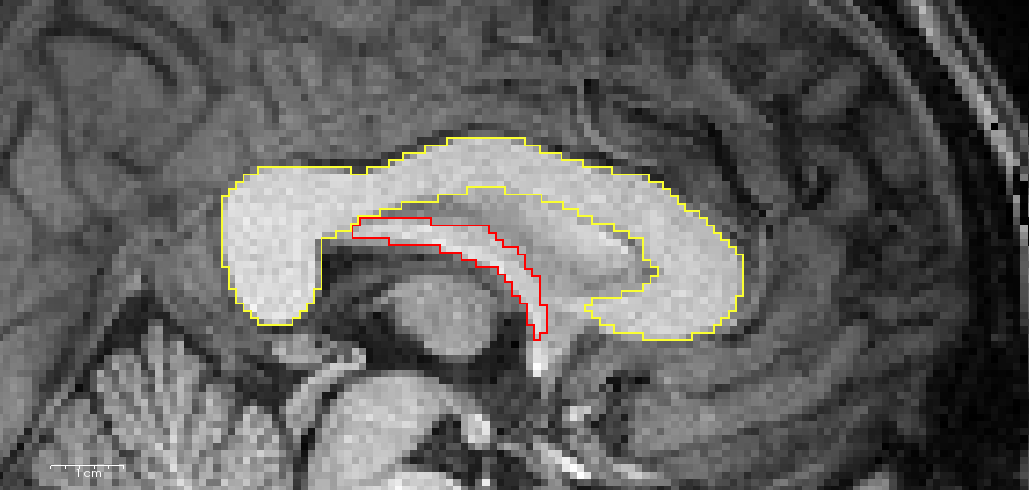}\hfill
    \includegraphics[width=\imagewidth]{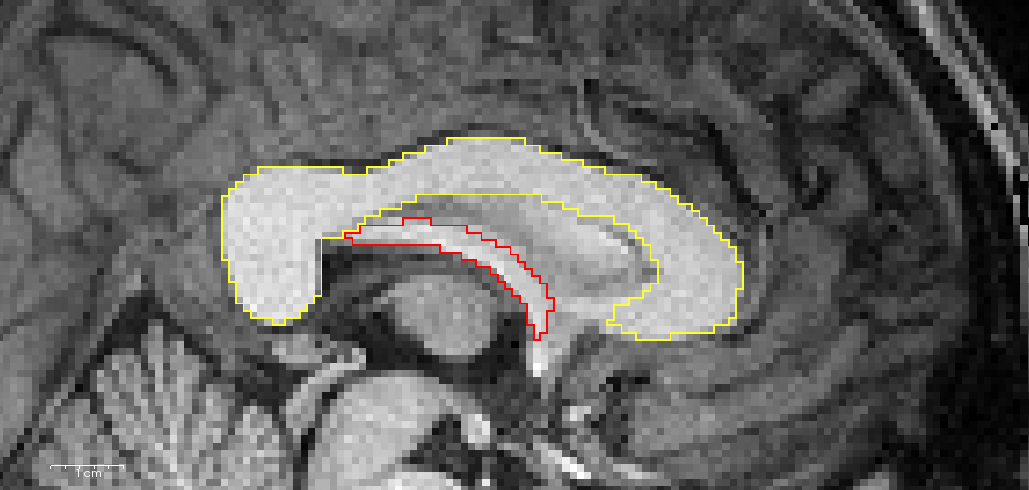} \\
    \vspace{1em} %

    \parbox[t]{\rowindicatorwidth}{\centering \textbf{B}}\hfill
    \includegraphics[width=\imagewidth]{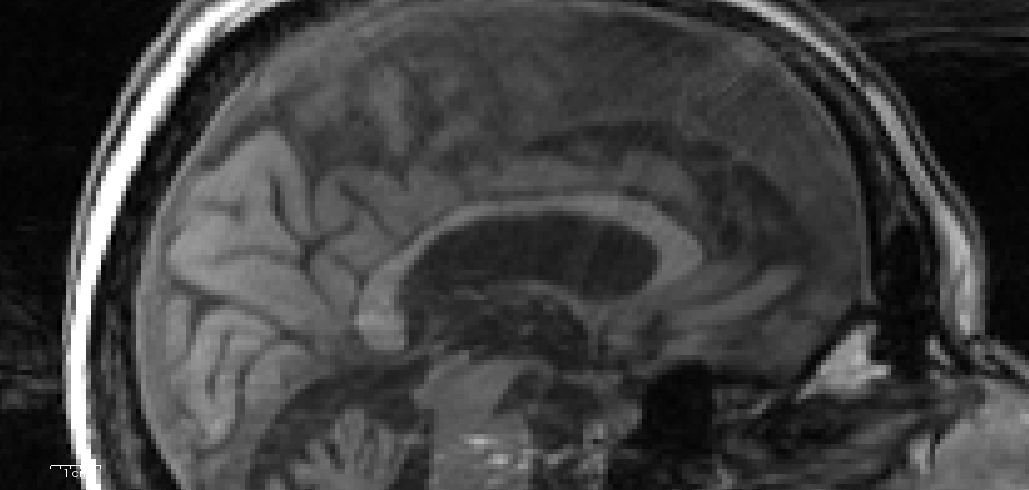}\hfill
    \includegraphics[width=\imagewidth]{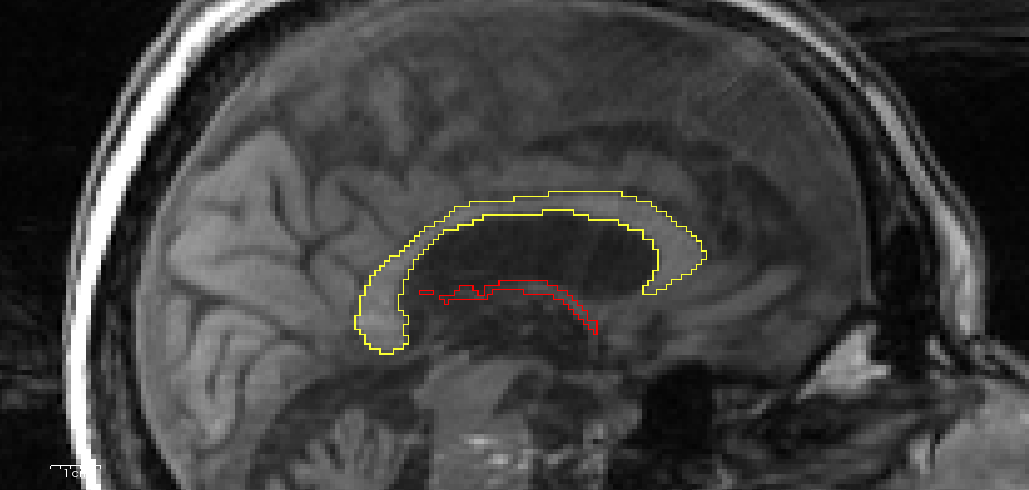}\hfill
    \includegraphics[width=\imagewidth]{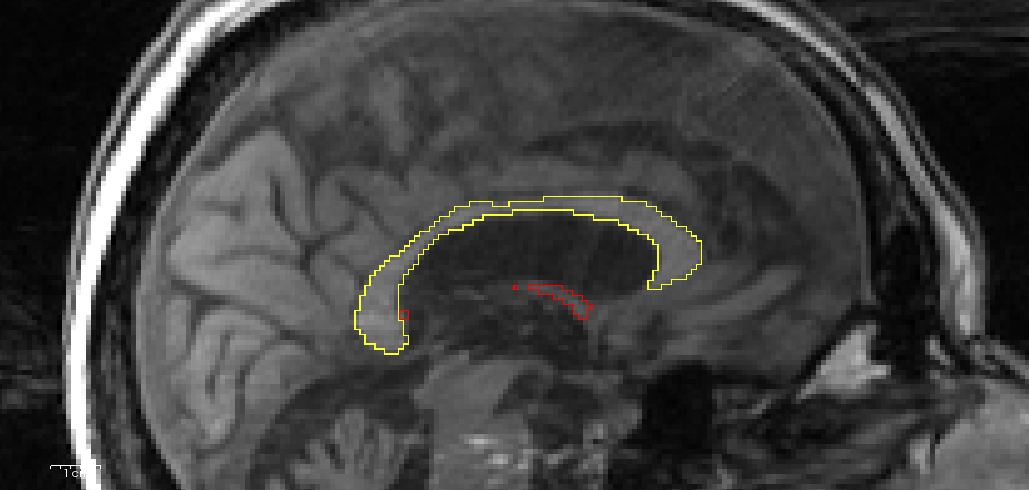} \\
    \vspace{1em}

    \parbox[t]{\rowindicatorwidth}{\centering \textbf{C}}\hfill
    \includegraphics[width=\imagewidth]{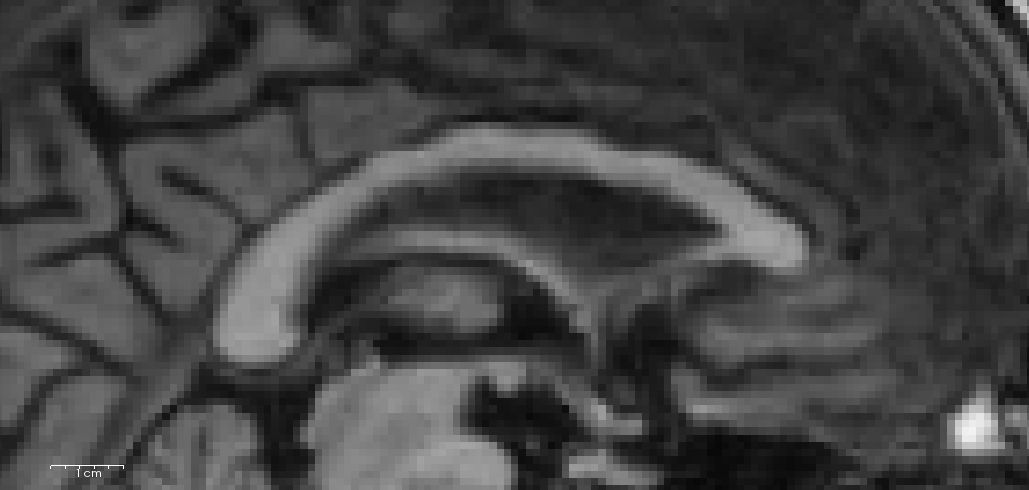}\hfill
    \includegraphics[width=\imagewidth]{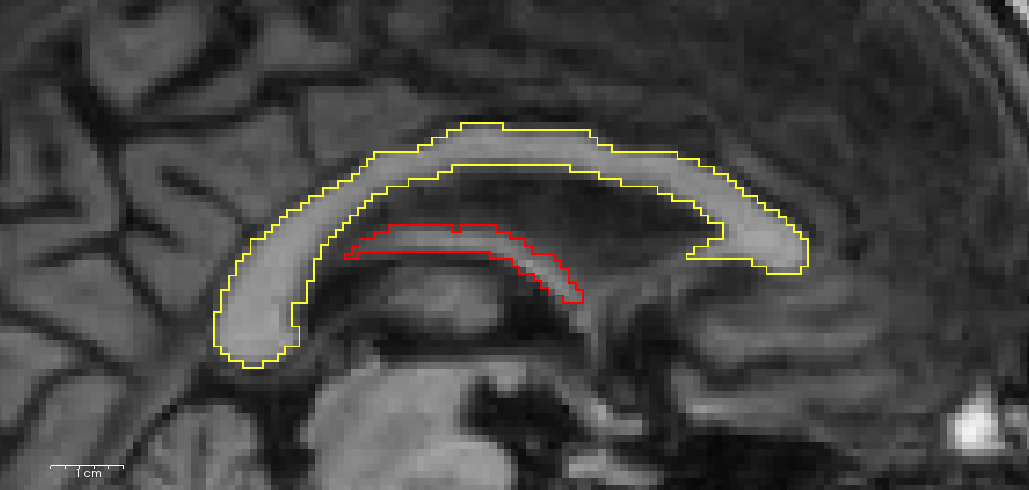}\hfill
    \includegraphics[width=\imagewidth]{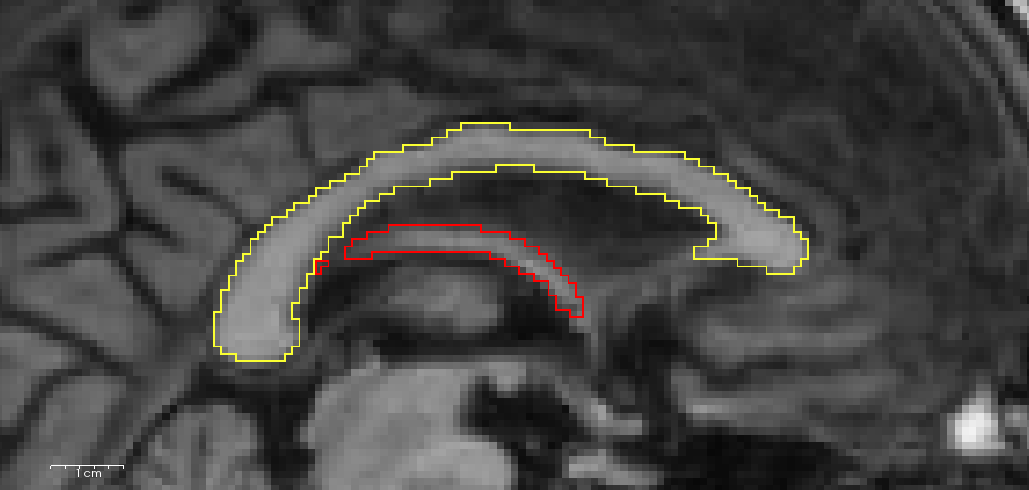} \\
    \vspace{1em}

    \parbox[t]{\rowindicatorwidth}{\centering \textbf{D}}\hfill
    \includegraphics[width=\imagewidth]{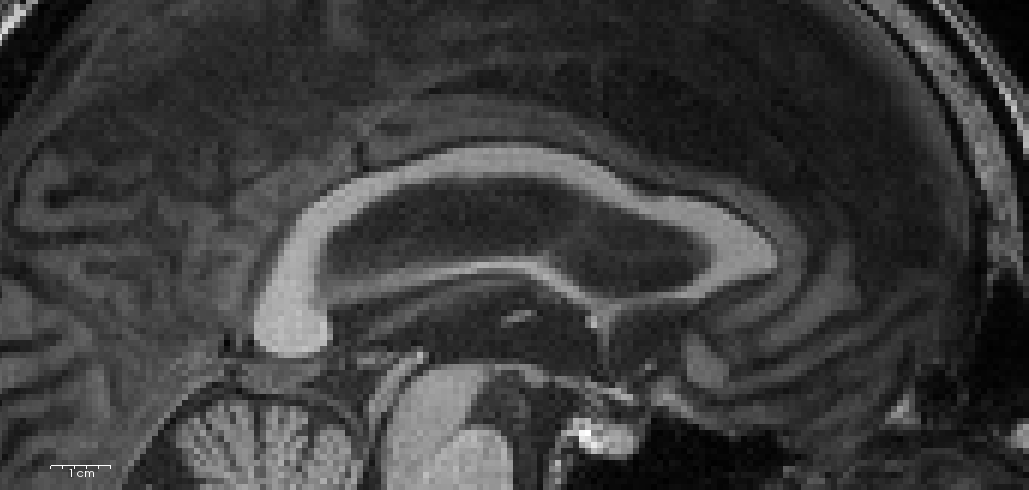}\hfill
    \includegraphics[width=\imagewidth]{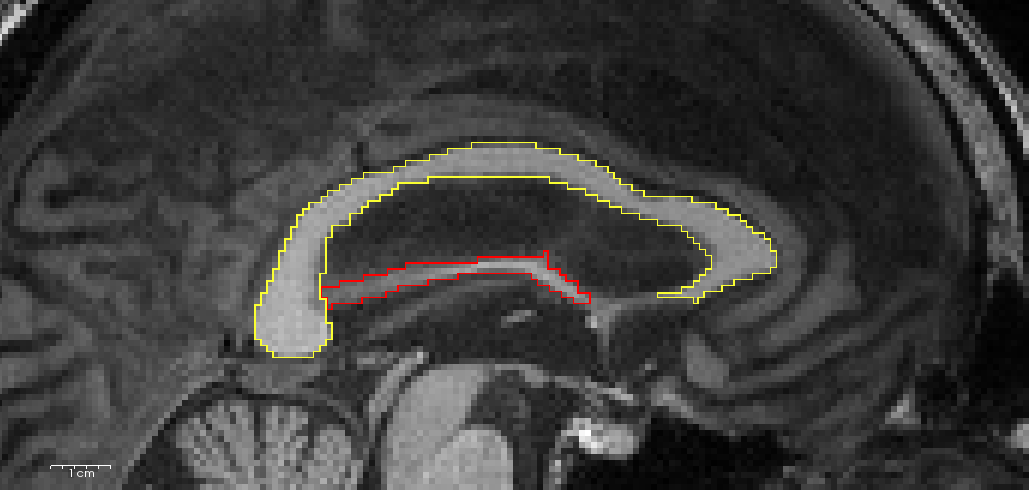}\hfill
    \includegraphics[width=\imagewidth]{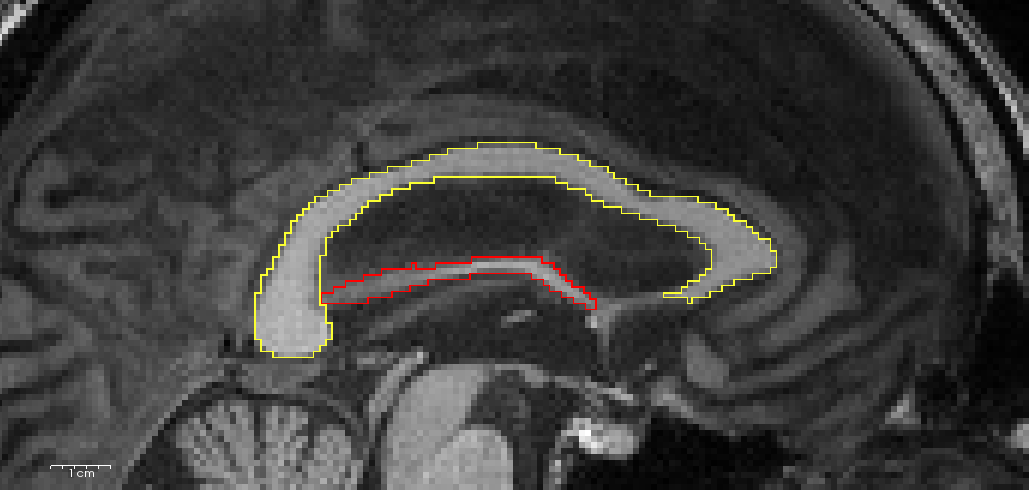} \\

    \caption{Example cases from the difficult test set with rater annotation and network prediction (unseen during training).}
    \label{fig:difficult_examples}
\end{figure*}

In Figure~\ref{fig:difficult_examples}, we show examples from the difficult test set along with the rater annotation and network output. In case A, the contrast of the corpus callosum and its surroundings is low (note that we use standardized automated contrast adjustment by Freeview for all figures). In case B, strong artifacts are present (see image background for a clear view of ringing artifacts). In cases B, C, and D, the corpus callosum is thin and deformed, making segmentation challenging.

\ifmulticol
\pagebreak
\else
\vspace{15cm}\hspace{2cm}\pagebreak
\fi

\printbibliography

\end{document}